\documentclass{article} 
\usepackage{iclr2026_conference,times} 

\usepackage{amsmath,amsfonts,bm}

\def\eqref#1{equation~\ref{#1}}

\def\1{\bm{1}}

\DeclareMathAlphabet{\mathsfit}{\encodingdefault}{\sfdefault}{m}{sl}
\SetMathAlphabet{\mathsfit}{bold}{\encodingdefault}{\sfdefault}{bx}{n}

\usepackage{hyperref}  
\usepackage{url}       
\usepackage{amsmath,amssymb}
\usepackage{booktabs}
\usepackage{graphicx} 
\usepackage[margin=1in]{geometry}
\usepackage{float}
\usepackage{subcaption}
\usepackage{multirow}
\usepackage{makecell}

\title{ECVL-ROUTER: Scenario-Aware Routing for Vision-Language Models}

\author{
Xin Tang$^{*1,2}$ \quad
Youfang Han$^{*2}$ \quad
Fangfei Gou$^{1}$ \quad
Wei Zhao$^{2}$ \quad
Xin Meng$^{2}$ \quad
Yang Yu$^{2}$ \\[0.8em]
\textbf{Jinguo Zhang}$^{2}$ \quad
\textbf{Yuanchun Shi}$^{2}$ \quad
\textbf{Yuntao Wang}$^{\dagger 1}$ \quad
\textbf{Tengxiang Zhang}$^{\dagger 2}$
\\[1.2em]
{\small $^{1}$Tsinghua University \qquad $^{2}$Goertek Inc.}
}
\iclrfinalcopy 
\begin{document}
\maketitle

\begin{abstract}

Vision-Language Models (VLMs) excel in diverse multimodal tasks. However, user requirements vary across scenarios, which can be categorized into fast response, high-quality output, and low energy consumption. Relying solely on large models deployed in the cloud for all queries often leads to high latency and energy cost, while small models deployed on edge devices are capable of handling simpler tasks with low latency and energy cost. To fully leverage the strengths of both large and small models, we propose ECVL-ROUTER, the first scenario-aware routing framework for VLMs. Our approach introduces a new routing strategy and evaluation metrics that dynamically select the appropriate model for each query based on user requirements, maximizing overall utility. We also construct a multimodal response-quality dataset tailored for router training and validate the approach through extensive experiments. Results show that our approach successfully routes over 80\% of queries to the small model while incurring less than 10\% drop in problem solving probability.

\end{abstract}

\section{Introduction}

Vision-Language Models (VLMs), which integrate visual and textual understanding, have become crucial components in a wide range of AI applications, from robotics control to user interface navigation \citep{zhang2024vision,shinde2025survey,li2024vision}. In practice, predominant deployment strategy relies heavily on powerful, cloud-hosted Large VLMs (LVLMs) to serve all user queries \citep{jang2025edge,navardi2025genai,zheng2025review}, which excel at complex reasoning but incur latency and energy costs \citep{fernandez2025energy,jegham2025hungry,charyyev2020latency} while underutilizing capable small VLMs (SVLMs) on edge devices \citep{sharshar2025vision,belcak2025small}. Moreover, a one-size-fits-all deployment strategy is suboptimal, as users increasingly expect systems that not only deliver high-quality responses but also adapt to diverse real-world scenarios with varying demands for latency, cost, and privacy.

To effectively integrate the strengths of both LVLMs and SVLMs, edge–cloud collaborative routing\citep{yuan2025local,hao2024hybrid} is a natural fits. At its core is a lightweight model router \citep{ding2024hybrid,ong2024routellm} that inspects each query and selects an appropriate VLM. However, a general router is insufficient, routing must be \textbf{scenario-aware}: behaviors vary across diverse application contexts and can be configured by users or automatically inferred by scenario detection algorithms \citep{fifty2023context,someki2025context}. Existing routers are often text-centric and optimize a fixed trade-off between cost and quality, failing to adapt to multimodal, scenario-aware user needs. For example, real-time games interaction prioritizes low latency, medical diagnostics emphasizes answer quality, and mobile assistants require low energy use and strong privacy \citep{asgari2025framework}. Therefore, an ideal model router should align with diverse user requirements and make the most appropriate routing decision in different scenarios.

In this work, we introduce \textbf{E}dge-\textbf{C}loud \textbf{V}ision-\textbf{L}anguage \textbf{Router} (\textbf{ECVL-ROUTER}), a novel, scenario-aware routing framework for VLMs. Motivated by heterogeneous real-world scenarios, we distill three primary user requirements: (1) fast response, (2) high-quality output, and (3) low energy consumption \& data privacy. Accordingly, our routing objective is to maximize the use of small edge models while meeting user satisfaction. To achieve this, we introduce the \emph{Minimal Expectation Score} (MES) to quantify the user's acceptable quality threshold in different scenarios. Guided by MES, the router prefers an SVLM whenever its predicted output meets the MES threshold and escalates to a larger cloud model otherwise. To implement this strategy, we design a lightweight routing classifier and construct a response quality dataset using an MES-based annotation policy to support its training. For evaluation, we propose the \emph{Routing Comprehensive Score} (RCS), integrating three key aspects: \emph{Average Problem-Solving Probability} (APSP), \emph{Cost Advantage} (CA), and \emph{Average Inference Latency} (AIL), which correspond directly to three core user requirements. Extensive experiments demonstrate that our routing strategy and framework delivers favorable trade-offs across quality, latency, and cost while adhering to scenario-specific user requirements.

The main contributions of this paper can be summarized as follows:
\begin{enumerate}
\item We propose a novel, scenario-aware routing strategy for VLMs and a new set of model routing evaluation metrics (Section \ref{sec:problem_formulation}). This strategy moves beyond the traditional cost-quality trade-off  by centering on the user's dynamic requirements to maximize the system's overall utility.

\item We design and implement ECVL-ROUTER, the first cloud-edge collaborative, scenario-aware routing framework for VLMs, which effectively combines the high performance of large cloud models with the low-cost advantages of small edge models (Section \ref{sec:ECVL-ROUTER}).

\item We construct the first response quality dataset for training and evaluating VLM routers (Section \ref{sec:dataset}). Through extensive experiments on this dataset, we have thoroughly validated the effectiveness of our ECVL-ROUTER framework (Section \ref{sec:experiments}). We also open source our dataset and framework for users to train their routers based on their specific requirements.
\footnote{The code is available at https://anonymous.4open.science/r/ECVL-Router-977D}
\end{enumerate}

\section{Related Work}

\subsection{SVLM vs.\ LVLM}
Vision-Language Models now underpin applications from accessibility assistance and UI navigation to robotics and scientific content creation \citep{zhang2024vision,shinde2025survey}. User requirements in these domains differ widely. Some applications, like real-time games or augmented reality, demand \textbf{fast response} where low latency is critical \citep{vasu2025fastvlm, liu2025perception}. Others, such as medical diagnostics, insist on \textbf{high-quality answers} where accuracy is paramount \citep{singhal2025toward,singhal2023large}. For applications on mobile or IoT devices, \textbf{low energy consumption and privacy constraints} are decisive factors \citep{chu2023mobilevlm,wang2025empowering}.

Cloud-hosted LVLMs (e.g., GPT-4o and Gemini 2.5 Pro) offer strong multimodal reasoning but impose round-trip latency and significant compute costs, making them ill-suited for latency-sensitive or privacy-constrained settings \citep{hurst2024gpt,team2023gemini,charyyev2020latency,fernandez2025energy}. In parallel, SVLMs have become increasingly capable on consumer hardware. For example, Google's Gemma 3 270M can run on a mobile phone with very low power consumption; internal tests showed that 25 conversations on a Pixel 9 Pro consumed only 0.75\% of the battery \citep{google_gemma3_270m_2025}. Meanwhile, Microsoft's Phi-4-multimodal, with only 5.6 billion parameters, runs on personal computers to efficiently handle tasks like OCR and chart understanding \citep{abouelenin2025phi}. Despite these advances, production stacks still over-rely on cloud LVLMs. This gap motivates scenario-aware edge–cloud designs that preferentially serve queries on-device and escalate only when quality requirements exceed local capacity.

\subsection{Model Routing}
\label{subsec:router}
From objective functions and decision paradigms, prior work falls into three types:

\noindent(1) \textbf{Outcome-optimal, non-predictive cascades.} 
These methods prioritize the quality of the final answer by allowing multiple model calls until a target quality threshold is met. A common strategy is \emph{cascading}: invoke models from low to high cost and stop once the response satisfies a predefined criterion. \emph{FrugalGPT} follows this paradigm \citep{chen2023frugalgpt}. \emph{AutoMix} first lets a small model self-evaluate its draft and escalates only if quality is predicted to be insufficient \citep{aggarwal2023automix}. While effective for quality, these approaches often incur significant extra latency due to repeated or parallel model evaluations.

\noindent(2) \textbf{Resource-optimal, predictive routing.} 
Unlike cascading methods, predictive routing seeks to choose the right model in a single shot, optimizing cost/latency by training a lightweight router that predicts the appropriate model before execution. \emph{Hybrid-LLM} uses a difficulty-aware router to decide between small and large models \citep{ding2024hybrid}. \emph{RouterLLM} provides a unified training framework for learning performance–cost aware selectors \citep{ong2024routellm}. NVIDIA's \emph{LLM-Router} leverages pretrained classifiers to dispatch by task type and complexity \citep{nvidia_2024}. Research also explores alternative designs, e.g., graph-based routing in \emph{GraphRouter} \citep{feng2024graphrouter} and tool-oriented modular routing in \emph{TO-Router} \citep{stripelis2024tensoropera}. However, many methods frame routing as predicting whether a small model can \emph{beat} a large model, rather than whether the small model is good enough for the scenario, a crucial distinction for practical deployments.

\noindent(3) \textbf{Compute-aware reasoning control in agentic systems.} 
Beyond model selection, routing principles appear in system architectures and agent workflows. Modern systems (e.g. GPT-5) use a real-time router to choose between instant response and deeper thinking for reasoning based on estimated task difficulty and explicit user intent \citep{openai_2025}. At the agent level, routing determines not only which LLM to use but also which tool or specialized sub-agent to activate within a complex system. \citep{wu2024autogen,yao2025toward}.

Despite progress, three limitations remain. (i) \textbf{Modality limitation:} prior routers target only text inputs and do not account for the unique challenges of multimodal (image–text) inputs required by VLMs. (ii) \textbf{Oversimplified routing strategy:} many method such as RouterLLM and Hybrid-LLM are driven by routing to a binary “small-vs-large wins” signal, by contrast, our router explicitly considers "good enough" performance of small models for the target scenario, improving edge utilization without unnecessary escalation. (iii) \textbf{Lack of scenario-aware user needs:} existing evaluations largely collapse to a single cost–quality trade-off and rely on cost/latency alone, overlooking that user requirements over answer quality, responsiveness, and resource usage vary across applications.

\section{Scenario-Aware VLM Routing}
\label{sec:problem_formulation}

This section formally defines our scenario-aware routing problem. We begin by defining the models and user-centric scenarios. Then we introduce the Minimal Expectation Score (MES) to quantify user satisfaction. we formally define the core routing problem and establish the decision rules. Finally, we propose a suite of evaluation metrics to assess the router's effectiveness across three scenarios.

\subsection{Routing Problem Definition}

\paragraph{Model definitions.}
Our framework considers two types of VLMs: (1) $M_{\mathrm{edge}}$, a lightweight SVLM deployed on consumer hardware such as laptops or mobile phones (typically $<\!10\mathrm{B}$ parameters); and (2) $M_{\mathrm{cloud}}$, a more powerful LVLM deployed on the cloud (typically $>\!10\mathrm{B}$ parameters). Each model $M$ maps a multimodal input----user query $Q$ and input data $I$ (text and images)----to a response: 
\begin{equation}
    M:\ \langle Q, I \rangle \mapsto \mathrm{Response} 
\end{equation}

\paragraph{Scenario definitions \& MES.}

We categorize user requirements into three primary scenarios: \textbf{(1) Fast Response} ($\mathsf{Speed}$), \textbf{(2) High-Quality Output} ($\mathsf{Quality}$), and \textbf{(3) Low Energy Consumption \& Data Privacy} ($\mathsf{Efficiency}$).

To quantify user requirements in our routing problem, we introduce the \textbf{Minimal Expectation Score (MES)}. The MES represents the lowest response-quality threshold a user is willing to accept in a given scenario. Let $\mathrm{Score}_{\mathrm{edge}}=\mathrm{Score}_{M_{\mathrm{edge}}}(Q,I)$ and $\mathrm{Score}_{\mathrm{cloud}}=\mathrm{Score}_{M_{\mathrm{cloud}}}(Q,I)$ denote response-quality scores (see Appendix~\ref{app:prs-rubric} for criteria). Any output with $\mathrm{Score}\ge\mathrm{MES}$ is deemed satisfactory; scores below $\mathrm{MES}$ are not. The MES effectively captures user needs across different scenarios; for instance, the $\mathsf{Quality}$ scenario typically demands a higher MES, while the $\mathsf{Speed}$ scenario may tolerate a lower one. Based on MES, we also construct a multimodal response-quality dataset for subsequent training and validation (see in Section \ref{sec:dataset}).

\paragraph{Routing problem.}
The core of our work is the routing problem, which aims to maximize the use of $M_{\mathrm{edge}}$ while satisfying the user's MES across different scenarios. Define a routing process $R$ that selects $M_R \in \{M_{\mathrm{edge}}, M_{\mathrm{cloud}}\}$. We say $M_{\mathrm{edge}}$ is \emph{competent} for $(Q,I)$ if the binary label $L=1$ under the following rule:
\begin{equation}
\label{eq:edge-competency}
L \;=\; \mathbb{I}\!\left[
\underbrace{\mathrm{Score}_{\mathrm{edge}} \,\ge\, \min\!\big\{\mathrm{Score}_{\mathrm{cloud}},\, \mathrm{MES}\big\}}_{\text{Case A: edge reaches the cloud-or-MES floor}}
\;\ \text{or}\ \;
\underbrace{\mathrm{Score}_{\mathrm{cloud}} \,<\, \mathrm{MES}}_{\text{Case B: cloud fails MES, avoid its cost}}
\right].
\end{equation}
\emph{Case A} routes to $M_{\mathrm{edge}}$ when it achieves at least the minimum of the cloud score and MES, which means $M_{\mathrm{edge}}$ is competent.  
\emph{Case B} covers failure regimes where even $M_{\mathrm{cloud}}$ cannot satisfy MES, which means the problem can't be solved. Using this rule, we construct training data $\mathcal{D}_{\mathrm{train}}=\{(Q_t, I_t, L_t)\}_{t=1}^{N}$, where $N$\label{parameter define} is the total number of samples in the training set, $(Q_t, I_t)$ is the $t$-th input query and data, $L_t$ is the $t$-th routing result.

\paragraph{Routing classifier and decision.}
The routing process is implemented by a \textbf{routing classifier}, which predicts the probability $p$ that $M_{\mathrm{edge}}$ is competent for a task: $p \;=\; P_{\theta}\!\left(L{=}1 \mid Q, I\right)$, where $\theta$ denotes model parameters learned by minimizing cross-entropy loss on $\mathcal{D}_{\text{train}}$. We use a decision threshold $\tau$ to convert $p$ into a routing decision:
\begin{equation}
\label{eq:routing-decision}
R(Q,I)=
\begin{cases}
M_{\mathrm{edge}}, & \text{if } p \,\ge\, \tau,\\[2pt]
M_{\mathrm{cloud}}, & \text{if } p \,<\, \tau.
\end{cases}
\end{equation}
Adjusting $\tau$ trades off quality, cost, and latency for different scenarios. (see in Section \ref{sec:ECVL-ROUTER})

\subsection{Evaluation Metrics}
\label{subsec:metrics}
To evaluate the scenario-aware router, we propose three metrics aligned with the core user requirements, along with a composite score. For the three, $(N,Q_t, I_t)$ share the same meaning with Section \ref{parameter define} and $M_{R_t}$ is the model selected by router $R$ for the $t$-th sample.

\paragraph{Average Problem-Solving Probability (APSP).}
Rate at which routed responses meet MES (aligns with $\mathsf{Quality}$):
\begin{equation}
\mathrm{APSP}(R) \;=\; \frac{1}{N}\sum_{t=1}^{N} \mathbb{I}\!\left\{\, \mathrm{Score}_{M_{R_t}}(Q_t, I_t) \,\ge\, \mathrm{MES} \,\right\}.
\end{equation}
A higher APSP indicates stronger problem-solving capability.

\paragraph{Cost Advantage (CA).}
Fraction of queries handled by $M_{\mathrm{edge}}$ (aligns with $\mathsf{Efficiency}$):
\begin{equation}
\mathrm{CA}(R) \;=\; \frac{1}{N}\sum_{t=1}^{N} \mathbb{I}\!\left\{\, M_{R_t} \,=\, M_{\mathrm{edge}} \,\right\}.
\end{equation}
A higher CA indicates better resource savings.

\paragraph{Average Inference Latency (AIL).}
Mean end-to-end latency (aligns with $\mathsf{Speed}$):
\begin{equation}
\mathrm{AIL}(R) \;=\; \frac{1}{N}\sum_{t=1}^{N} \mathrm{latency}\!\left(M_{R_t}(Q_t, I_t)\right).
\end{equation}
A lower AIL indicates faster responses.

\paragraph{Routing Comprehensive Score (RCS).}
A weighted aggregate (higher is better):
\begin{equation}
\mathrm{RCS}(R) \;=\; \alpha\, \mathrm{APSP}(R) \;+\; \beta\, \mathrm{CA}(R) \;-\; \gamma\, \mathrm{AIL}(R).
\end{equation}
The hyperparameters $\alpha, \beta, \gamma$ can be tuned based on scenario priorities. A high APSP often leads to a lower CA and a higher AIL, as more tasks are routed to the LVLM. The RCS enables balanced evaluation. In our experiments (see in Sec \ref{sec:experiments}), we use three parameterizations to simulate different user priorities:
\[
\text{RCS1}_{\mathsf{(Quality)}}:\ (\alpha,\beta,\gamma)=(1.2,\,0.1,\,0.001),\quad
\text{RCS2}_{\mathsf{(Efficiency)}}:\ (1.0,\,0.12,\,0.001),\quad
\text{RCS3}_{\mathsf{(Speed)}}:\ (1.0,\,0.1,\,0.0015).
\]

\section{Methodology}

\subsection{Response Score Dataset}
\label{sec:dataset}

\paragraph{Dataset construction.}
Guided by the scenario-aware MES defined in Sec.\ref{sec:problem_formulation}, we construct the Response Score Dataset (RSD) tailored for VLM routing. Our goal is to efficiently estimate model response quality across diverse tasks via large-scale automated annotation. Concretely, we first select a representative set of \textbf{8} VLMs and prompt them to generate responses on \textbf{7} public benchmarks that cover a broad range of task types and difficulty levels (see in Appendix~\ref{app:rsd-models}). Next, we adopt the \emph{LLM-as-a-Judge} way, which has shown to correlate strongly with human ratings in prior work \citep{zheng2023judging,gu2024survey} and use GPT-4o as the \emph{LLM Response Judge (LRJ)}. For each model response, the LRJ assigns a unified $score \in [1,10]$ under a consistent rubric. The score evaluates the helpfulness, relevance, accuracy of the response (see in Appendix~\ref{app:prs-rubric} for details). 

\paragraph{Human gold labels}
To ensure the reliability of automated scoring, we rigorously validated LRJ annotations against human gold labels. We recruited five human experts to independently score a stratified random sample of 200 instances. LRJ scores show strong alignment with the mean human rating (Pearson correlation $r>0.85$), confirming its effectiveness and reliability as an automated annotator (see Appendix~\ref{app:gold-validation} for detailes). 

In summary, the Response Score Dataset (RSD) contains approximately \textbf{22k} image-text instances, each with response quality scores and inference times for \textbf{8} VLMs. Furthermore, we analyze the distribution of response quality and inference time, as well as the performance of different models across each benchmark (see Appendix~\ref{app:dataset-stats} for full analysis). This helps us define the difficulty of each benchmark and the capabilities of different models. The distribution result also supports the hypothesis that larger models typically take more time to provide higher-quality responses.\footnote{The construction of RSD—including model inference, LRJ, and human gold validation—incurred USD~\$1000 cost.}

\subsection{ECVL-ROUTER}
\label{sec:ECVL-ROUTER}
\begin{figure}
    \centering
    \includegraphics[width=0.75\linewidth]{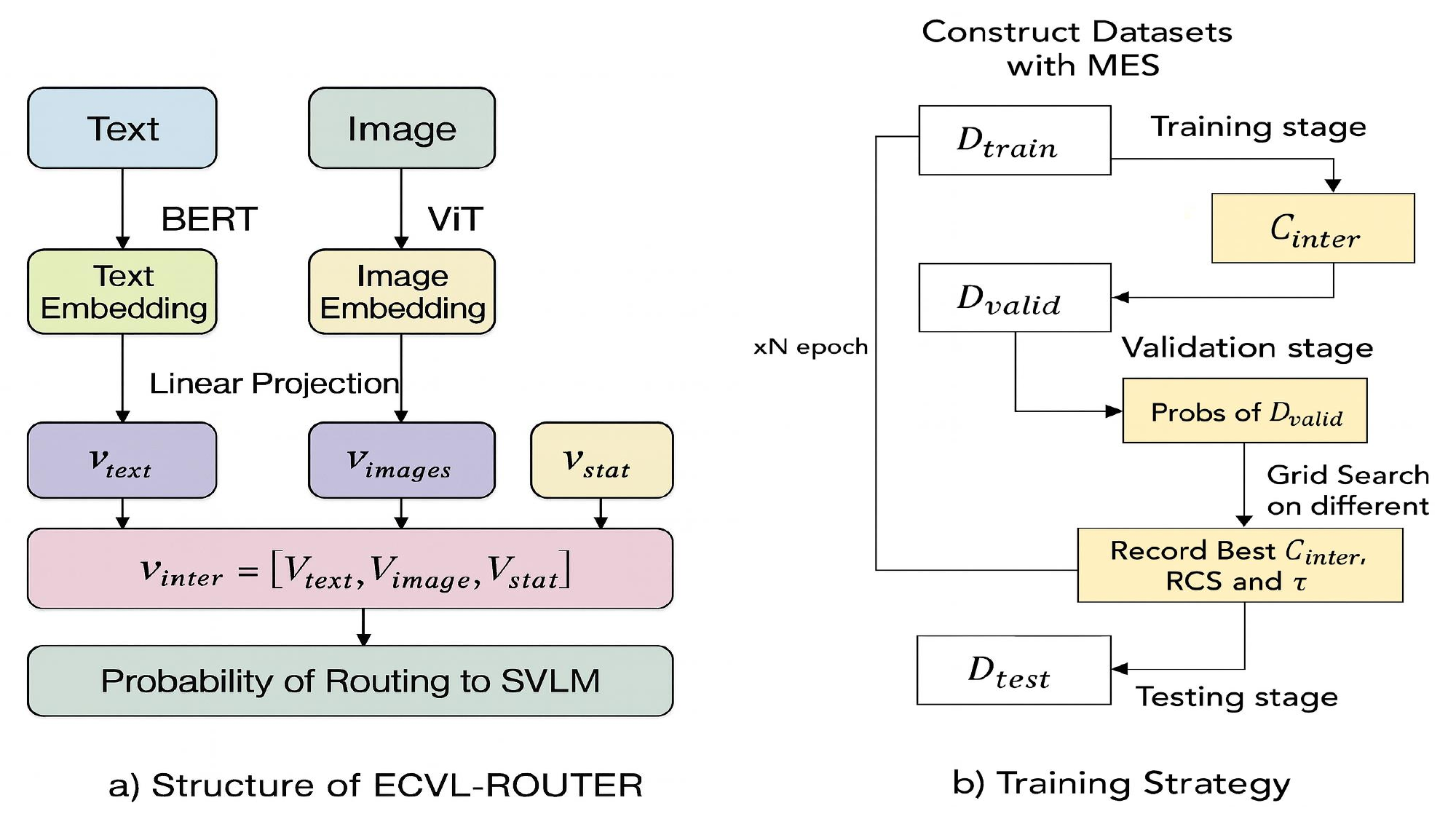}
    \caption{(a) Overall Structure and (b)Training Strategy of ECVL-ROUTER }
    \label{fig:model}
\end{figure}

We propose ECVL-ROUTER, a transformer-based routing framework for Vision-language models. ECVL-ROUTER uses a Transformer classifier to process heterogeneous inputs (e.g. text and images). On the training side, \textsc{ECVL-ROUTER} selects the decision threshold \(\tau\) via validation-set analysis, yielding a threshold tailored to each model pair, MES setting, and application scenario. The overall architecture and training pipeline of \textsc{ECVL-ROUTER} are shown in Fig \ref{fig:model}.

\subsubsection{Model Architecture}

\paragraph{Input encoding.}
We obtain modality-specific embeddings for text and images. For text, we use a pretrained encoder (e.g.BERT\_BASE) to produce $\mathbf{e}_{\text{text}}$. For images, we adopt a standard vision encoder (e.g., ViT) to obtain $\mathbf{e}_{\text{image}}$. Each embedding $\mathbf{e}_m \in \mathbb{R}^{k_m}$ is projected into a shared $d$-dimensional space via a linear layer, i.e., $\mathbf{v}_m=\mathbf{W}_m \mathbf{e}_m+\mathbf{b}_m$, where $\mathbf{W}_m \in \mathbb{R}^{d \times k_m}$ and $\mathbf{b}_m \in \mathbb{R}^{d}$, with $m \in \{\text{text},\text{image}\}$, and $k_m$ is the dimensionality of the vector obtained by applying a linear projection to modality $m$.
.
\paragraph{Modalities fusion.}
To capture input complexity in addition to semantics, we compute lightweight statistics from raw inputs (e.g., word count, special-character and numeric-token counts for text; width/height and color-channel indicators for images). These statistics are linearly embedded as $\mathbf{v}_{\text{stat}} \in \mathbb{R}^{d}$. We then form the intermediate representation by concatenation: $\mathbf{v}_{\text{inter}}=[\mathbf{v}_{\text{text}};\mathbf{v}_{\text{image}};\mathbf{v}_{\text{stat}}]$.
\paragraph{Routing classifier.}
ECVL-ROUTER employs a lightweight Transformer encoder (2 layers, hidden size 256, 4 attention heads, FFN size 512, dropout 0.3) followed by a linear classifier. Given $\mathbf{v}_{\text{inter}}$ as input, the classifier outputs the probability that the edge-side SVLM can meet the scenario-dependent MES, modeled as $p_{\theta}(\text{SVLM}\mid q,\mathbf{x})=\sigma(\mathbf{w}^{\top}\mathbf{h}_{\text{out}}+b)$, where $\theta$ denotes all model parameters, $\mathbf{h}_{\text{out}}$ is the final encoder representation, and $\sigma(\cdot)$ is the sigmoid function. At inference time, we route to the SVLM if $p_{\theta}\ge \tau$ and otherwise escalate to the cloud-side LVLM, with $\tau \in (0,1)$ fixed by the validation procedure described above.

\subsubsection{Training Strategy}

\paragraph{Data construction.}
We first structure the original corpora into triples \(D=\{(Q,I,L)\}\) following Sec.~\ref{sec:problem_formulation}. Here, \(Q\) denotes the user query, \(I\) denotes the associated input (images and text), and \(L\in\{0,1\}\) is the \emph{edge-competency} label computed by ~\eqref{eq:edge-competency}. The dataset is then randomly split into the training set \(D_{\text{train}}\), validation set \(D_{\text{valid}}\), and test set \(D_{\text{test}}\) with a ratio of \(6{:}2{:}2\), while preserving the distribution of sources and labels.
\paragraph{Training stage.}
At each epoch, an intermediate classifier \(C_{\text{inter}}\) is trained on \(D_{\text{train}}\). The classifier outputs \(p=P_{\theta}(L{=}1\mid Q,I)\), the probability that \(M_{\mathrm{edge}}\) is competent for \((Q,I)\) as defined in Sec.~\ref{sec:problem_formulation}. Given the ground-truth label \(y=L\in\{0,1\}\), parameters are updated by minimizing the binary cross-entropy \(\mathcal{L}=-\big[y\log p+(1-y)\log(1-p)\big]\). We use Adam with an initial learning rate \(10^{-3}\) and the OneCycleLR schedule\citep{smith2019super} to leverage super-convergence. The model is trained for a total of \(50\) epochs with a batch size of \(64\).
\paragraph{Validation stage.}
During validation, we perform a grid search over the decision threshold \(\tau\in[0,1]\) with a step of \(0.05\) to maximize the RCS on \(D_{\text{valid}}\). For each epoch, we evaluate \(\mathrm{RCS}(D_{\text{valid}};\tau)\) for all candidate \(\tau\), record the best pair \((\tau^\star,\mathrm{RCS}^\star)\), and finally retain the \(C_{\text{inter}}\) achieving the highest \(\mathrm{RCS}^\star\) together with its \(\tau^\star\). This procedure adapts the router to scenario-specific requirements and generalizes well to the test set (see in Fig.\ref{fig:align}).

\section{Experiments}
\label{sec:experiments}

\subsection{Experimental setup}
\paragraph{Base VLMs \& Comparison Methods.}
We run all experiments offline on an NVIDIA A800 (80\,GB) and an Intel\textsuperscript{\textregistered} Xeon\textsuperscript{\textregistered} Platinum 8250C (128 cores). To avoid API latency and reduce cost, the main study uses open source VLMs from OpenGVLab at three scales: InternVL3-38B, InternVL3-8B, and InternVL2.5-1B. These form three model pairs, and we include additional pairs for generalization tests(see in Appendix \ref{app:adapt}). Due to there are no previous works in VLM router, following text-only router work \citep{ong2024routellm,ding2024hybrid}, we compare against \textbf{All-at-Large}, \textbf{All-at-Small}: route all queries to large/small model deployed on cloud/edge, and Matrix Factorization(\textbf{MF}). We also evaluate Gradient Boosted Decision Trees (\textbf{GBDT}, 100 trees) and a multilayer perceptron (\textbf{MLP}) with three hidden layers. For the MLP, input/output dimensions and the optimizer learning rate match those of ECVL-ROUTER.

\paragraph{Dataset \&  Evaluation Metrics.}
We utilize the Response Score Dataset, where samples are labeled based on a MES of 6. A sample is positive when the SVLM reaches $\mathrm{Score}_{edge}\ge6$. The dataset is split into training, validation, and test sets at 60\%, 20\%, and 20\%, respectively, ensuring stratified label distributions. During evaluation, we report the per-scenario metrics \textbf{APSP},\textbf{CA} and \textbf{AIL}, along with their composite scores \textbf{RCS1}, \textbf{RCS2}, and \textbf{RCS3} for the three different scenarios.

\begin{table}[h]
\centering
\small
\caption{Model router performance on three model pairs. APSP, CA, AIL are the metrics in RCS1. RCS1/2/3 are composite scores for \emph{Quality}, \emph{Efficiency}, and \emph{Speed} scenarios.}
\label{tab:router}
\renewcommand{\arraystretch}{1.1}
\begin{tabular}{c l c c c c c c}
\toprule
\multirow{2}{*}{\textbf{Model Pair}} & \multirow{2}{*}{\textbf{Model Router}} &
\multicolumn{3}{c}{\textbf{Evaluation metrics(RCS1)}} &
\multicolumn{3}{c}{\textbf{Composite Scores}} \\
 & & \textbf{APSP}~$\uparrow$ & \textbf{CA}~$\uparrow$ & \textbf{AIL [s]}~$\downarrow$ & \textbf{RCS1}~$\uparrow$ & \textbf{RCS2}~$\uparrow$ & \textbf{RCS3}~$\uparrow$ \\
\midrule
\multirow{6}{*}{\makecell{InternVL\\38B/1B}}
 & \textbf{ECVL-ROUTER} & 0.506 & 0.824 & 4.53 & \textbf{0.685} & \textbf{0.601} & \textbf{0.582} \\
 & GBDT                 & 0.518 & 0.631 & 5.38 & 0.680 & 0.596 & 0.577 \\
 & MLP                  & 0.515 & 0.645 & 4.41 & 0.678 & 0.594 & 0.575 \\
 & MF                   & 0.503 & 0.439 & 4.49 & 0.643 & 0.551 & 0.540 \\
 & All-at-Large                  & \textbf{0.549} & 0.000 & 7.44 & 0.652 & 0.542 & 0.538 \\
 & All-at-Small                  & 0.456 & \textbf{1.000} & \textbf{0.94} & 0.646 & 0.575 & 0.554 \\
\cmidrule(lr){1-8}
\multirow{6}{*}{\makecell{InternVL\\8B/1B}}
 & \textbf{ECVL-ROUTER} & 0.483 & 0.910 & 1.34 & \textbf{0.669} & \textbf{0.591} & \textbf{0.572} \\
 & GBDT                 & 0.478 & 0.941 & 1.29 & 0.666 & 0.589 & 0.570 \\
 & MLP                  & 0.485 & 0.873 & 1.24 & 0.668 & 0.589 & 0.570 \\
 & MF                   & 0.469 & 0.800 & 1.18 & 0.642 & 0.564 & 0.547 \\
 & All-at-Large                  & \textbf{0.529} & 0.000 & 1.63 & 0.633 & 0.527 & 0.527 \\
 & All-at-Small                  & 0.456 & \textbf{1.000} & \textbf{0.94} & 0.646 & 0.575 & 0.554 \\
\cmidrule(lr){1-8}
\multirow{6}{*}{\makecell{InternVL\\38B/8B}}
 & \textbf{ECVL-ROUTER} & 0.533 & 0.982 & 1.77 & \textbf{0.736} & \textbf{0.649} & \textbf{0.629} \\
 & GBDT                 & 0.534 & 0.965 & 1.86 & 0.735 & 0.648 & 0.628 \\
 & MLP                  & 0.534 & 0.887 & 2.30 & 0.727 & 0.638 & 0.619 \\
 & MF                   & 0.529 & \textbf{1.000} & \textbf{1.63} & 0.733 & 0.647 & 0.627 \\
 & All-at-Large                  & \textbf{0.549} & 0.000 & 7.44 & 0.652 & 0.542 & 0.538 \\
 & All-at-Small                  & 0.529 & 1.000 & 1.63 & 0.733 & 0.647 & 0.627 \\
\bottomrule
\end{tabular}
\end{table}

\subsection{Router Performance Results}
\label{subsec:results}

\autoref{tab:router} shows the performance of our ECVL-ROUTER against comparison methods across three model pairs. Based on this, we can draw the following observations:

\paragraph{Obs 1: ECVL-ROUTER achieves optimal performance across all scenarios.}
The experimental results show that our proposed ECVL-ROUTER consistently achieves the highest composite scores on all three model pairs across the three scenarios. Beyond these in-domain results, we also conduct the cross-domain experiments in Appendix \ref{app:generate}.

\paragraph{Obs 2: ECVL-ROUTER achieve high edge(small) model utilization with minimal quality loss and lower latency.}
While maintaining high-quality responses, ECVL-ROUTER substantially increases the utilization of the small model and significantly reduces response latency. For instance, compared to the All-at-Large baseline on the 38B/1B pair, ECVL-ROUTER routes 82\% of queries to the SVLM. This is achieved with only a minor drop in the APSP of less than 8\% (0.549 $\rightarrow$ 0.506), while delivering a significant 39.1\% reduction in AIL.

\subsection{Analysis of Scenario-aware Hyperparameters ($\tau$ and MES)}
\label{subsec:hyperparams}

Our router is controlled by two hyperparameters: the decision threshold $\tau$ and MES. We pick $\tau$ via validation grid search for each scenario and validate its robustness in further experiments in Appendix \ref{app:generate}. MES is application-defined and proxies task difficulty (higher MES $\Rightarrow$ harder). We analyzes the impact of varying $\tau$ and MES on the router's performance. For all other experiments, MES is set to 6 by default. From figure \ref{fig:sensi-tau} and \ref{fig:sensi-mes}, we can find that:

\begin{figure}[h]
    \centering
    \includegraphics[width=\linewidth]{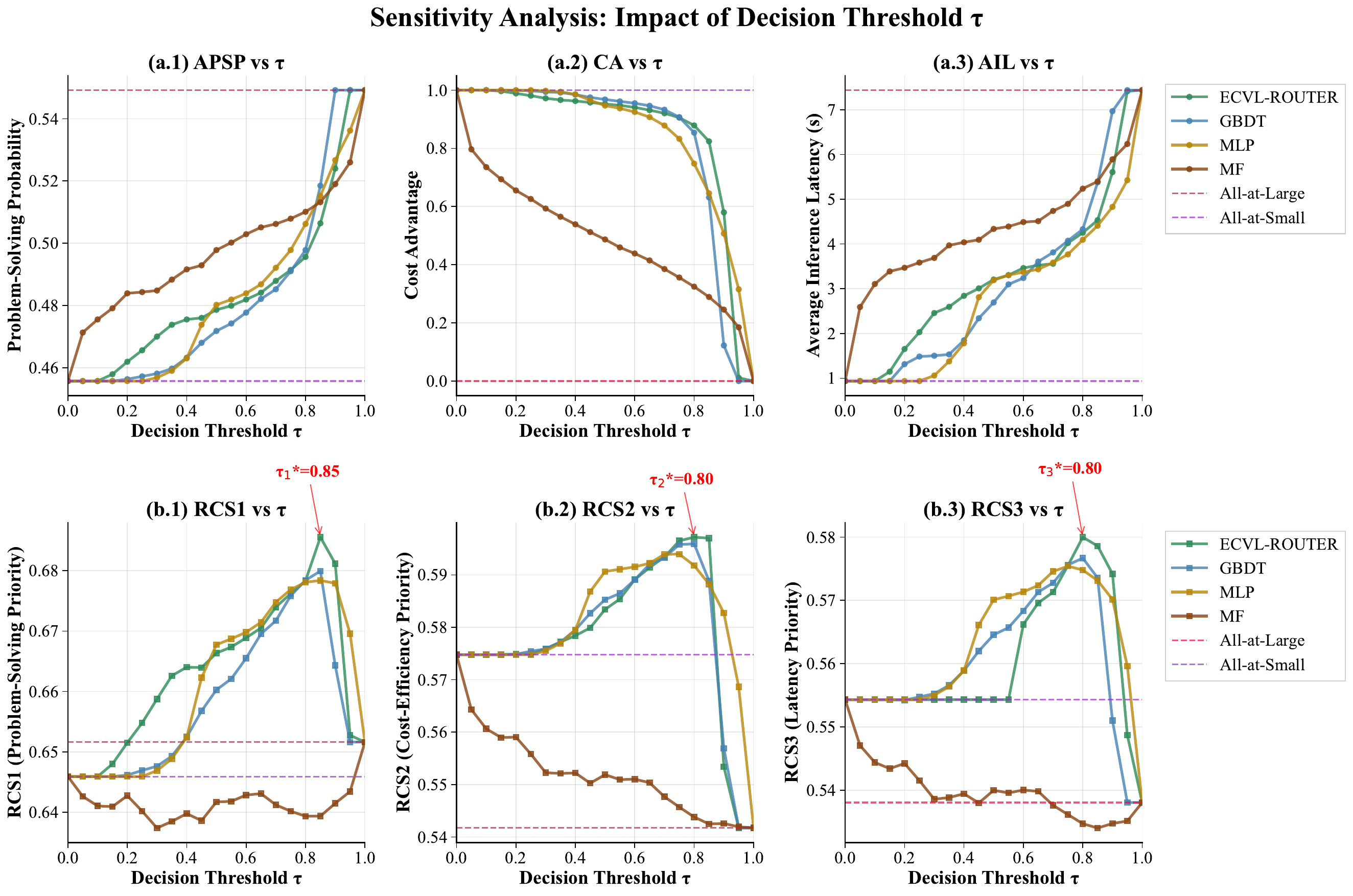}
    \caption{Impact of the decision threshold $\tau$. It illustrates how the performance metrics for each model router (ECVL-ROUTER, MLP, GBDT, MF) change with different values of $\tau$ for the InternVL-38B/1B pair at MES=6.}
    \label{fig:sensi-tau}
\end{figure}

\paragraph{Obs 3: The decision threshold $\tau$ governs the cost-quality trade-off, with an optimal value $\tau^\star$ that is scenario-dependent and lies within [0.7, 0.9].}

In figure \ref{fig:sensi-tau}(a), as $\tau$ increases, fewer queries go to the SVLM: CA $\downarrow$, while escalation to the LVLM raises APSP $\uparrow$ and AIL $\uparrow$. This captures the core trade-off: higher quality costs more latency/energy; speed efficiency implies some quality drop. By finding the optimal $\tau$ on the validation set, we can identify the best value for a specific scenario. In figure \ref{fig:sensi-tau}(b), across RCS1/2/3, most methods (except MF) show unimodal curves peaking in $0.6$–$0.9$ rather than at $\tau{=}0.5$, whcih means $\tau$ has optimal value and is differed by scenarios. As seen in the red annotations of figure~\ref{fig:sensi-tau}(b), scenarios prioritizing quality (RCS1) favor a higher optimal $\tau^\star$, while those focused on low cost and speed (RCS2, RCS3) benefit from a lower optimal $\tau^\star$. Our router leads across most $\tau$, with peak performance often at higher $\tau$ (0.7–0.9). In contrast, MLP and GBDT vary smoothly, whereas MF is limited by linear factorization and misses rich image–text interactions.

\paragraph{Obs 4: ECVL-ROUTER delivers the largest gains at medium difficulty task (MES $\in$ [5, 7]).}
 When MES$<5$, tasks are easy and the SVLM already succeeds; when MES$>8$, both models often fail, leaving little room for routing. Concretely, In MES$=5$–$7$, ECVL-ROUTER shows the largest margin over alternatives (e.g., average RCS gain = +0.7\% in figure\ref{fig:sensi-mes}(b),), because the small-large model capability gap is most informative: the router accurately escalates only those queries the SVLM cannot solve, maximizing edge usage without undue quality loss. And in \ref{fig:sensi-mes}(a), as MES increases, APSP decreases monotonically, while CA (first $\downarrow$, then $\uparrow$) and AIL (first $\uparrow$, then $\downarrow$) are non-monotonic because the rule defaults to the SVLM when neither model meets MES; thus at very high MES,  the proportion of queries handled by the SVLM rises again. A detailed analysis of these cases is provided in Appendix~\ref{app:unexpect}.


\subsection{Router Latency Analysis \& Ablation Experiment}

We measure end-to-end inference latency on the test set for the InternVL family and for each component of ECVL-ROUTER. From Table~\ref{tab:model_latency_comparison} and \ref{tab:ablation_study}, we have the following observations:

\paragraph{Obs 5: Routing latency of ECVL-ROUTER is negligible for user experience.}
One ECVL-ROUTER pass takes 0.0159\,s, i.e., 1.7\% of InternVL2.5-1B, 0.97\% of InternVL3-8B, and 0.21\% of InternVL3-38B. Within the router, the ViT encoder dominates (93.7\%), while fusion is near zero (0.06\%). In practice, routing overhead is amortized by the subsequent VLM call and does not affect user-perceived latency.

\begin{table}[H]
\centering
\scriptsize
\setlength{\tabcolsep}{3pt}
\caption{Comparison of Inference Latency Across Different Models.}
\label{tab:model_latency_comparison}
\resizebox{\linewidth}{!}{
\begin{tabular}{lcccccccc}
\toprule
 & Router-Bert & Router-ViT & Router-Stat & Router-Fusion & Router-All & InternVL2.5-1B & InternVL3-8B & InternVL3-38B \\
\midrule
Latency [s]
& \(7.5\times 10^{-4}\)
& 0.0149
& \(2.3\times 10^{-4}\)
& \(8.8\times 10^{-6}\)
& \textbf{0.0159}
& 0.9359
& 1.6332
& 7.4391 \\
 &
 &
 &
 &
 & (1.7\% of 1B)
 & (baseline)
 & (1.75$\times$ 1B)
 & (7.95$\times$ 1B) \\
\bottomrule
\end{tabular}}
\end{table}

\paragraph{Obs 6: Visual modality is the dominant driver in VLM routing.}
Our ablation studies, summarized in Table~\ref{tab:ablation_study}, reveal that visual features are not just beneficial but are the primary signal guiding the routing decision for VLM pairs. Removing any branch degrades all scenarios while dropping image hurts most (e.g., $\Delta$RCS2 = $-0.022$, $\Delta$RCS3 = $-0.017$); compared with removing text ($\Delta$RCS2 = $-0.017$) and statistics yields smaller. Furthermore, among single-branch variants, image-only $>$ text-only on all composites and is competitive with statistics-only.

\begin{table}[h]
\centering
\caption{Ablation of ECVL-ROUTER's Components (Text, Image, Statistics) on the InternVL 38B/1B Model Pair.}
\label{tab:ablation_study}
\begin{tabular}{lccccccc}
\toprule
\textbf{Method} & \textbf{APSP}~$\uparrow$ & \textbf{CA}~$\uparrow$ & \textbf{AIL[s]}~$\downarrow$ & \textbf{RCS1}~$\uparrow$ & \textbf{RCS2}~$\uparrow$ & \textbf{RCS3}~$\uparrow$ \\
\midrule
ECVL-ROUTER (Full) & 0.5064 & 0.8241 & 4.5331 & \textbf{0.6855} & \textbf{0.6008} & \textbf{0.5820} \\
w.o. Text & 0.5079 & 0.6720 & 4.9529 & 0.6718 & 0.5836 & 0.5767 \\
w.o. Image & \textbf{0.5174} & 0.5496 & 4.7378 & 0.6711 & 0.5786 & 0.5653 \\
w.o. Statistic & 0.5150 & 0.6567 & 5.1550 & 0.6785 & 0.5886 & 0.5729 \\
only Text & 0.5152 & 0.5710 & 5.0925 & 0.6702 & 0.5786 & 0.5647 \\
only Image & \textbf{0.5174} & 0.5807 & 4.9666 & 0.6740 & 0.5821 & 0.5680 \\
only Statistic & 0.4894 & \textbf{0.9032} & \textbf{4.4539} & 0.6732 & 0.5933 & 0.5730 \\
\midrule
Random & 0.4980 & 0.5000 & 4.1653 & 0.6434 & 0.5538 & 0.5418 \\
All-at-Large & \textbf{0.5492} & 0.0000 & 7.4391 & 0.6516 & 0.5418 & 0.5380 \\
All-at-Small & 0.4557 & \textbf{1.0000} & \textbf{0.9359} & 0.6459 & 0.5748 & 0.5543 \\
\bottomrule
\end{tabular}
\end{table}

\subsection{Alternate Routing Strategy \& Metrics}
\label{subsec:alt-routing}
We reimplement two widely used routing paradigms: \emph{win-hard} and \emph{win-soft}, following the spirit of Hybrid LLM and RouteLLM.\footnote{All methods use our unified LRJ 1--10 scale for comparison}
\autoref{tab:alt-routing-offline} reports per-scenario metrics(APSP/CA/AIL),  the quality composite RCS1, router accuracy (ACC), the RouteLLM metric PGR, HybridLLM metric BARTScore, and aggregate Token/Time saving.

\begin{table}[h]
\centering
\small
\setlength{\tabcolsep}{7pt}
\caption{Offline comparison of routing strategies under a unified label/metric protocol. \textbf{Bold}=best; \underline{underline}=second-best. \emph{win-hard}: direct comparison: $\mathrm{Score}_{\mathrm{edge}}\ge\mathrm{Score}_{\mathrm{cloud}}$. \emph{win-soft}$(k)$: Score-offset variant favoring SVLM by $k\in1,2$; during labeling and decision the SVLM is judged against $\mathrm{Score}_{\mathrm{edge}}+k\ge\mathrm{Score}_{\mathrm{cloud}}$. }
\label{tab:alt-routing-offline}
\begin{tabular}{lcccc}
\toprule
\textbf{Metric} & \textbf{Proposed} & \textbf{win-hard} & \textbf{win-soft (1)} & \textbf{win-soft (2)} \\
\midrule
APSP~\(\uparrow\)      & \underline{0.5064} & \textbf{0.5250} & 0.5012 & 0.4784 \\
CA~\(\uparrow\)        & \underline{0.8241} & 0.4008 & 0.7601 & \textbf{0.8609} \\
AIL~[s]~\(\downarrow\) & \underline{4.5331} & 6.8286 & 5.1018 & \textbf{3.5520} \\
RCS1~\(\uparrow\)      & \textbf{0.6855} & 0.6633 & \underline{0.6723} & 0.6566 \\
ACC~\(\uparrow\)       & \textbf{0.8408} & \underline{0.8159} & 0.8113 & 0.8011 \\
PGR~\(\uparrow\)      & \underline{0.7720} & \textbf{0.9486} & 0.6001 & 0.3645 \\
BARTScore~\(\uparrow\) & \underline{-4.8154} & \textbf{-4.6750} & -4.8869 & -5.0040 \\
\midrule
Token saving~\(\uparrow\) & \underline{28{,}269} & 9{,}341 & 25{,}292 & \textbf{33{,}136} \\
Time saving~\(\uparrow\)  & \underline{12{,}544} & 4{,}858 & 10{,}324 & \textbf{14{,}717} \\
\bottomrule
\end{tabular}
\end{table}

\paragraph{Obs 7: Scenario-Aware routing outperforms Single-Objective strategies.} Our method ranks highest on the composite \(\mathrm{RCS1}{=}0.6855\) and on ACC (0.8408), and second on APSP/CA/AIL/PGR/BARTScore. It achieves the highest \(\mathrm{RCS1}{=}0.6855\) with APSP close to \emph{win-hard} (0.5064 vs.\ 0.525), while routing more to the edge (CA \(+\!0.4233\)\(\uparrow\)) and cutting latency (AIL \(-2.30\)s\(\downarrow\)).
Against \emph{win-soft}(1/2) (MES-shifted), it improves APSP/CA/AIL jointly (e.g., vs.\ soft(1): APSP \(+\!0.0052\)\(\uparrow\), CA \(+\!0.0640\)\(\uparrow\), AIL \(-0.57\)s\(\downarrow\)). Notably, PGR is highest for \emph{win-hard}, consistent with RouteLLM’s cost–performance emphasis, while our RCS objective aligns better with scenario-aware utility that balances quality, edge utilization, and latency. Hybrid LLM reports up to \(40\%\) fewer large-model calls at no quality drop under its thresholds; under the same unified LRJ regime, \textsc{ECVL-ROUTER} achieves stronger composite utility while maintaining high edge coverage. Furthermore,
to assess real-world adaptability, we run an online study with \(n{=}5\) participants over \(200\) image–text queries under two scenarios---Quality-first and Speed-first---using the same router trained offline. Full online study task analyses appear in Appendix~\ref{app:online}.

\section{Discussion}
\label{sec:discussion}

\paragraph{Point 1: Scenario-aware routing objective to align with user needs.}
We redefine routing as \emph{satisficing} against a Minimal Expectation Score (MES): a response is ``good enough'' if it meets the user-specified minimum. This aligns routing with real application needs and maximizes the utility of on-device SVLMs by sending easy queries locally while escalating only when necessary. The approach is most effective when there is a clear capability gap between the SVLM and LVLM; when the SVLM is too weak (e.g., $<\!500$M parameters) or the two models are similar in ability, gains from routing are limited. These findings offer actionable guidance for edge--cloud collaboration: set acceptance thresholds by user needs to minimize compute while maintaining satisfaction.

\paragraph{Point 2: Router flexibility and deployment practice.}
\textsc{ECVL-ROUTER} instantiates the above strategy with user-definable \emph{MES} and threshold $\tau$. Practitioners can construct training data with LRJ labels at a chosen MES and train with the composite metric \textsc{RCS} using weights $(\alpha,\beta,\gamma)$ over APSP/CA/AIL (see Sec.~\ref{subsec:metrics}) to encode scenario preferences; the optimal $\tau$ is then selected on a validation set. In practice, we recommend curating in-domain data and tuning $(\alpha,\beta,\gamma)$ to the target scenario, as cross-domain generalization is modest. Relative to All-at-Large processing, our approach substantially increases edge model utilization with only small satisfaction drops, cutting both latency and cloud cost.

\section{Conclusion}
\label{sec:conclusion}
We introduced \textsc{ECVL-ROUTER}, a scenario-aware routing framework for vision--language models that reframes routing as meeting a user-defined Minimal Expectation Score (MES). By optimizing the Routing Comprehensive Score (RCS)---which balances quality (APSP), efficiency (CA), and latency (AIL), our method maximizes on-device small model usage while preserving response quality. Trained on the \textsc{RSD} dataset ($\sim$22k image--text instances), \textsc{ECVL-ROUTER} consistently outperforms baselines across multiple model pairs, routing over 80\% of queries to the edge with less than 10\% drop in APSP and substantial end-to-end latency reductions. The decision threshold $\tau$ enables scenario-specific trade-offs; ablations show visual features dominate routing efficacy, and the router’s overhead ($\approx$0.016,s) is negligible. The approach works best when edge and cloud capabilities differ markedly. Future work will focus on enhancing cross-domain generalization and extending the framework to other modalities.

\clearpage

\section*{Ethics Statement}
This work develops a scenario-aware routing method for vision–language systems using publicly available datasets and open-source/base models. No personally identifiable information was collected or released; all data use complies with original licenses/terms. We label responses via an LLM-as-a-Judge rubric and perform human sanity checks to assess agreement and common biases (e.g., over-verbosity, modality imbalance), applying mitigation such as a dimensioned rubric, separation of judge/model, and conservative decision thresholds. Potential risks include unsafe escalation policies, amplification of dataset bias, and privacy leakage in cloud calls. We recommend deployment safeguards: scenario-appropriate MES settings, content-safety filters, rate/permission controls for remote inference, and clear documentation of intended use and known limitations. Code, configs, and evaluation scripts are released to support auditing and community oversight.

\section*{Reproducibility Statement}
We release an anonymous repository containing code, configs, and scripts for dataset preparation, training, and evaluation, together with a readme that enumerates dependencies and exact commands. We document model pairs, routing architecture, hyperparameters, random seeds, and the 60/20/20 stratified splits; evaluation follows APSP/CA/AIL with composite RCS. Threshold selection uses a validation grid over $\tau$; we provide the sweep script and the best setting per scenario. Hardware/software details (e.g., NVIDIA A800 80\,GB; Intel Xeon) and environment files are included. All reported tables/figures can be reproduced by running the provided pipelines; ablations and latency measurement scripts are also supplied. For reviewers, we include an anonymized code link and archival artifact with cached intermediate results to reduce compute and facilitate verification.

\section*{LLM Usage}
We used ChatGPT (OpenAI; Aug--Sep 2025 access window) for grammar checking and minor phrasing, and AI-assisted coding tools (e.g., Cursor) for refactoring boilerplate and editor suggestions during system development.All LLM outputs were reviewed by authors. We verified factual claims, math, code logic, citations, and figures; any errors were corrected by humans. We did not upload non-public data or PII to third-party services. Prompts contained only de-identified text or synthetic/task data.


\bibliography{refs}
\bibliographystyle{iclr2026_conference}

LLMs are not authors. The authors take full responsibility for all content, results, and any errors.

\appendix

\clearpage

\section{Response Score Dataset}
\subsection{Models and Dataset Composition}
\label{app:rsd-models}

\paragraph{Models.}
We evaluate a pool of VLMs that spans both large cloud models $M_{\mathrm{cloud}}$ (LVLM) and small edge models $M_{\mathrm{edge}}$ (SVLM), following the definitions in \S\ref{sec:problem_formulation}.

\begin{table}[H]
\caption{Model roster by category.}
\centering
\small
\begin{tabular}{ll}
\toprule
Category & Models \\
\midrule
Large (LVLM) & Gemma 3-27B;\; InternVL3-38B \\
Small (SVLM) & InternVL2\_5-2B;\; InternVL2\_5-1B;\; SmolVLM-256M;\\
             & InternVL3-8B;\; Phi-4-Multimodal-5.6B;\; Qwen2.5-VL-7B \\
\bottomrule
\end{tabular}
\label{tab:model_roster}
\end{table}

\paragraph{Dataset composition.}
RSD covers diverse tasks and difficulty levels to support MES-based scenario-aware routing. Table~\ref{tab:datasets} reports \emph{Core Task Type}, \emph{Key Abilities}, dataset \emph{Volume (k)}, and \emph{Difficulty} (three levels: Easy/Medium/Hard). The total annotated instances are about $\sim 22$k model--instance pairs, aligned with the discussion in the main text.

\begin{table}[H]
\caption{Task coverage of RSD training/eval data. Volumes (in thousands) sum to $\sim$22k overall. Difficulty uses three discrete levels (Easy/Medium/Hard) to coarsely stratify instance hardness for MES analysis.}
\centering
\small
\begin{tabular}{lllcc}
\toprule
Dataset & Core Task Type & Key Abilities & Volume (k) & Difficulty \\
\midrule
WildVision & Real-world VQA & open-ended reasoning, contextual understanding & 0.5 & Easy \\
ChartQA & Chart QA & Structured extraction, logic/arithmetic & 2.5 & Easy \\
GQA & Compositional VQA & Spatial reasoning, multi-step inference & 12.0 & Medium \\
VizWiz & Blind-assistance VQA & Noise robustness, answerability & 4.3 & Medium \\
MMVet & Composite benchmark & Recognition/OCR/knowledge/spatial/math & 0.22 & Medium \\
MMMU-Pro & Professional hard VQA & Domain knowledge, deep reasoning & 1.73 & Hard \\
MMStar & Leak-resistant eval & Fine-grained/counterfactual, vision reliance & 1.5 & Hard \\
\midrule
\multicolumn{3}{r}{\textbf{Total}} & \textbf{22.7} & \textemdash \\
\bottomrule
\end{tabular}
\label{tab:datasets}
\end{table}

\subsection{LLM-as-a-Judge Prompt \& Biases Control}
\label{app:prs-rubric}

\paragraph{Scoring Prompt.}
We adopt an \emph{LLM-as-a-Judge} (LRJ) procedure to assign a unified $score\in[1,10]$ per model--instance pair. The LRJ (GPT-4/4o in our setup) evaluates \emph{accuracy, completeness, relevance, and level of detail} under a consistent rubric; the resulting 1--10 scale aligns with the scenario-specific MES threshold $\tau$ used by the router in \S\ref{sec:problem_formulation}. Human--LRJ agreement on a stratified sample demonstrates high correlation. We use the following template to elicit a scalar score followed by a short rationale:

\begin{small}
\begin{verbatim}
SCORING_TEMPLATE_1_10 = """You are a helpful and precise assistant for checking
the quality of multimodal AI responses.

[Question]
{question}

[Reference Answer]
{reference}

[Model Answer]
{prediction}

[System]
We would like to request your feedback on the performance of the AI assistant
in response to the user question displayed above.

Please rate the helpfulness, relevance, accuracy, level of details of the
response. The assistant receives an overall score on a scale of 1 to 10,
where a higher score indicates better overall performance.

Scoring Guidelines:
1: Completely unable to answer the task, content is completely unrelated
   to the question, or refuses to answer
2: Attempts to answer but severely deviates from the topic, contains obvious
   understanding errors, completely incorrect information
3: Partially understands the question but answer quality is extremely poor,
   contains multiple major errors, illogical
4: Basically understands the question but answer is inaccurate, contains
   some important errors, lacks key information
5: Understands the question and provides relevant answer, but insufficient
   accuracy, obvious defects
6: Answer is basically correct, can execute the task but with average
   effect, passing line, some minor errors
7: Answer is accurate and has certain logic, can execute the task well,
   good quality
8: Answer is accurate, detailed and logically clear, almost no errors,
   excellent quality
9: Answer is complete, basically consistent with reference answer, high
   accuracy, clear and complete expression
10: Answer not only meets reference answer requirements, but also considers
    more aspects, more comprehensive, exceeds expectations

Please first output a single line containing only one value indicating the
score for the model. In the subsequent line, please provide a comprehensive
explanation of your evaluation, focusing on the accuracy, completeness,
logic, and relevance of the response.

Score: """
\end{verbatim}
\end{small}
\paragraph{Bias Control}
Our scoring design explicitly addresses common biases:

\begin{itemize}
  \item \textbf{Self-enhancement bias.} \emph{Risk}: a model favors its own outputs. \emph{Mitigation}: the evaluator (LRJ) is strictly disjoint from the evaluated models; we use GPT-4/4o solely as the judge.
  \item \textbf{Knowledge limitation bias.} \emph{Risk}: judge hallucination or gaps hurt accuracy. \emph{Mitigation}: provide a \emph{Reference Answer} (ground truth) and a structured rubric focusing on correctness/relevance, reducing reliance on the judge’s prior.
  \item \textbf{Multimodal-specific biases.} \emph{Risks}: (i) over-reliance on salient visual cues; (ii) inconsistency handling when text and image conflict. \emph{Mitigation}: use a multimodal-capable judge (GPT-4o) and require dimension-wise assessment (accuracy, completeness, logic, relevance) to encourage cross-modal consistency checks.
\end{itemize}

\subsection{Gold-Standard Validation (Human Expert)}
\label{app:gold-validation}
We randomly sample 200 items from RSD and collect eight-model responses per item (\(200\times 8=1{,}600\) LRJ–human pairs). Five anonymized experts (Expert1–5) independently score each item; their mean serves as the gold standard. LRJ aligns closely with humans: Pearson \(r=0.8805\), Spearman \(\rho=0.8349\) (both \(p<10^{-4}\)). Agreement is high (79.83\% within 1 point; 90.75\% within 2 points), and mean inter-rater correlation is 0.8615. A paired \(t\)-test shows a small bias (human–LRJ \(\Delta\approx-0.375\); \(t=-23.93\); \(p<10^{-4}\)). Per-evaluator correlations are reported in Table~\ref{tab:expert_corr}.

\begin{table}[h]
\caption{Per-evaluator correlations between LRJ and individual experts on the 200-sample validation (\(1{,}600\) pairs). Experts are anonymized as \emph{Expert1–5}.}
\centering
\small
\begin{tabular}{lc}
\toprule
\textbf{Evaluator} & \textbf{Pearson \(r\)} \\
\midrule
Expert1 & 0.883 \\
Expert2 & 0.930 \\
Expert3 & 0.894 \\
Expert4 & 0.866 \\
Expert5 & 0.903 \\
\bottomrule
\end{tabular}
\label{tab:expert_corr}
\end{table}

\subsection{Score and Latency Distributions}
\label{app:dataset-stats}
\subsubsection{Score Distribution}
\begin{figure}[h]
    \centering
    \includegraphics[width=0.7\linewidth]{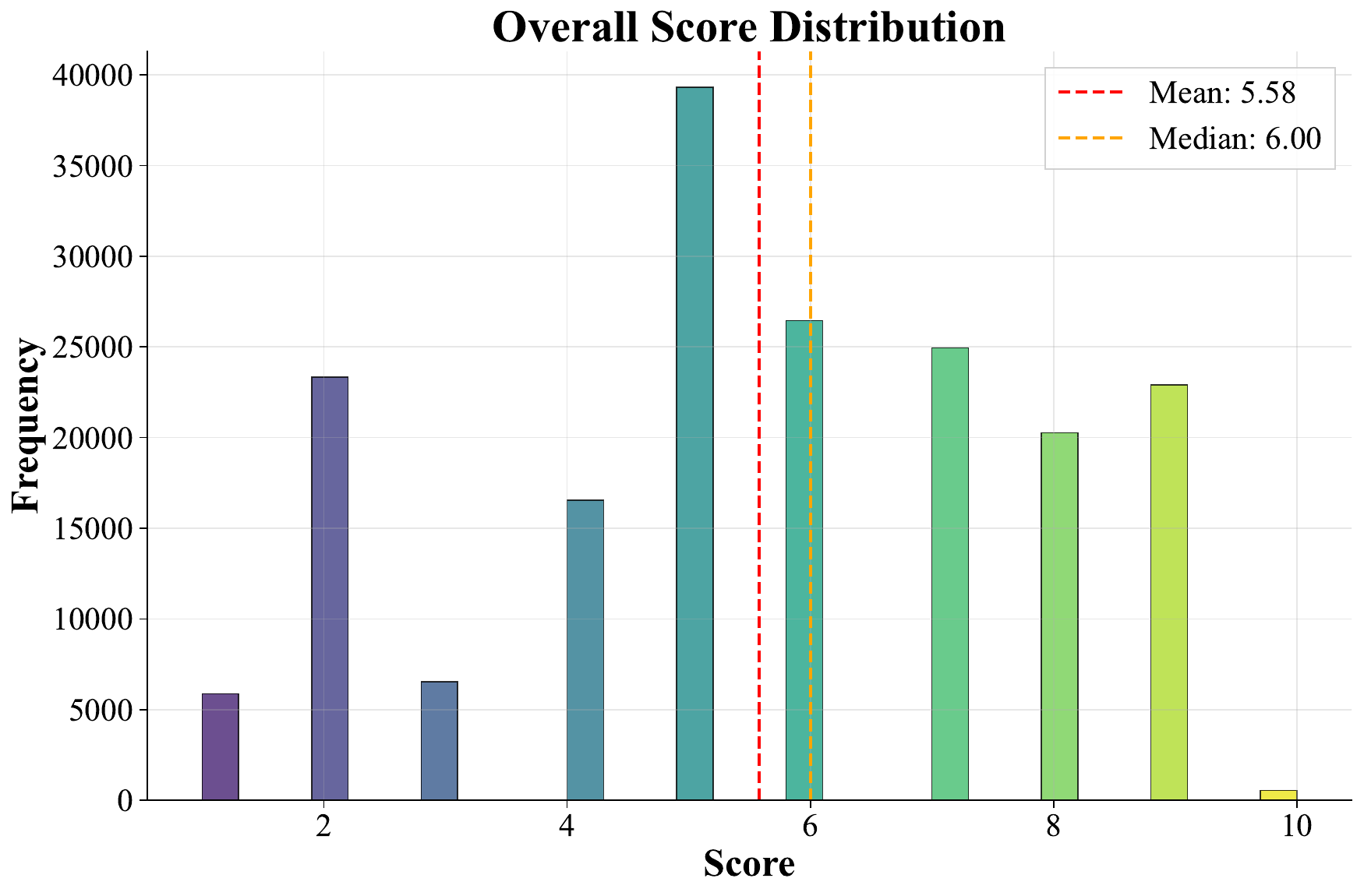}
    \caption{Overall score histogram across all models and datasets in RSD ($\sim$22k pairs). Dashed lines mark the mean ($\approx 5.58$) and median ($\approx 6.00$).}
    \label{fig:rsd-score-overall}
\end{figure}

\begin{figure}[h]
    \centering
    \begin{subfigure}{0.49\linewidth}
        \centering
        \includegraphics[width=\linewidth]{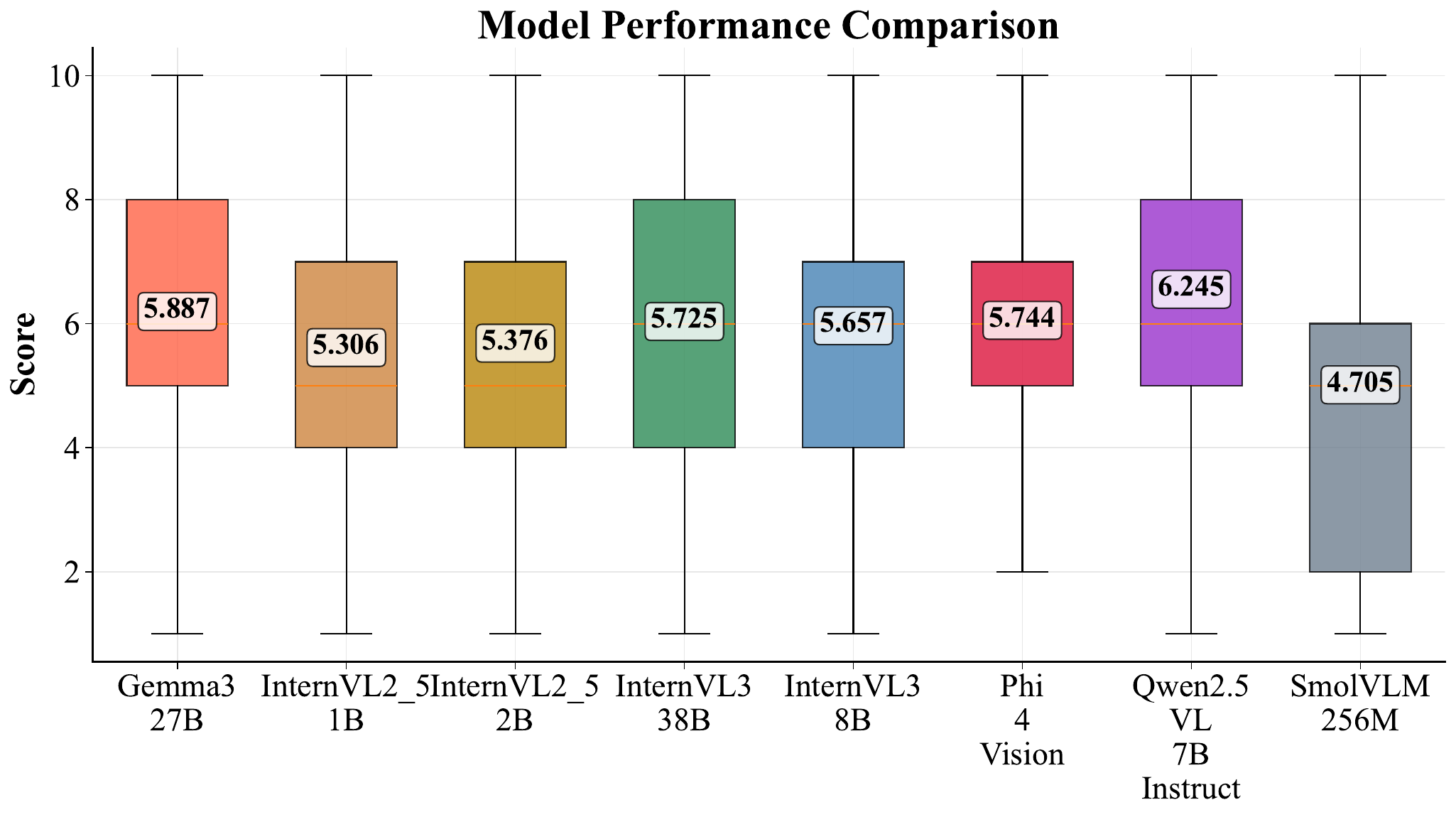}
        \caption{Dataset difficulty ranking (mean $\pm$ SE). Higher mean $\Rightarrow$ easier.}
        \label{fig:rsd-difficulty}
    \end{subfigure}\hfill
    \begin{subfigure}{0.49\linewidth}
        \centering
        \includegraphics[width=\linewidth]{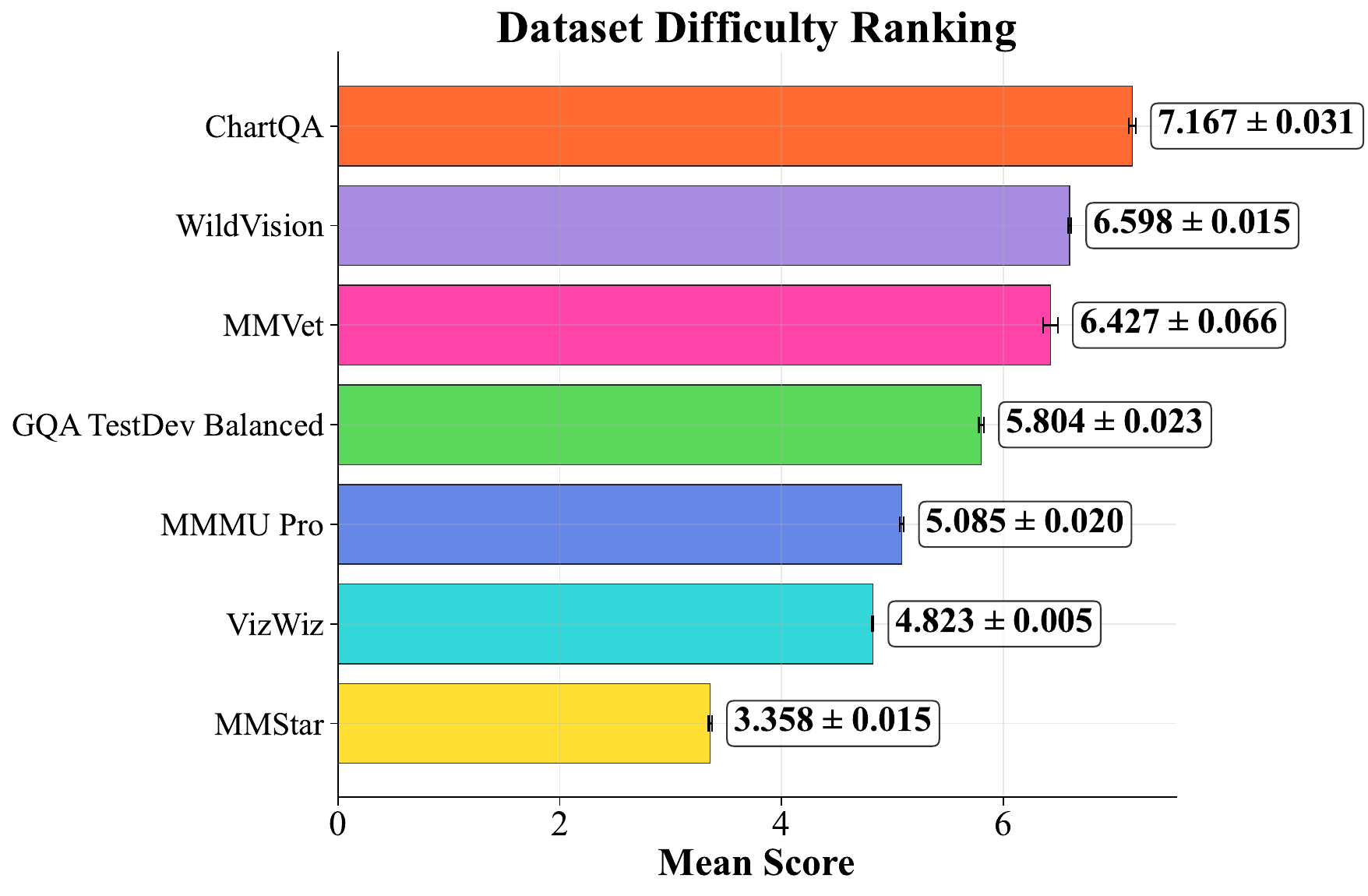}
        \caption{Per-model score distributions (boxplots with annotated means).}
        \label{fig:rsd-model-box}
    \end{subfigure}
    \caption{Score breakdown in RSD: (a) difficulty by dataset; (b) performance by model across the same instances.}
    \label{fig:rsd-score-breakdown}
\end{figure}

RSD aggregates $\sim$22k model–instance pairs spanning seven VLMs (SVLM$\leftrightarrow$LVLM) and eight benchmarks (Appendix~\ref{app:rsd-models}, Table~\ref{tab:datasets}). Figure~\ref{fig:rsd-score-overall} shows an approximately spread 1–10 score distribution with a mean of $\approx 5.6$ and median $\approx 6.0$, consistent with our MES-aligned rubric. Figure~\ref{fig:rsd-score-breakdown} further decomposes scores: (a) dataset-level means reveal a clear difficulty gradient (MMStar hardest; ChartQA easiest), and (b) per-model boxplots indicate overlapping performance across scales, with several SVLMs competitive with LVLMs. These statistics ground our scenario-aware routing analysis; latency distributions are reported in the next subsection.

\begin{figure}[h]
    \centering
    \includegraphics[width=1\linewidth]{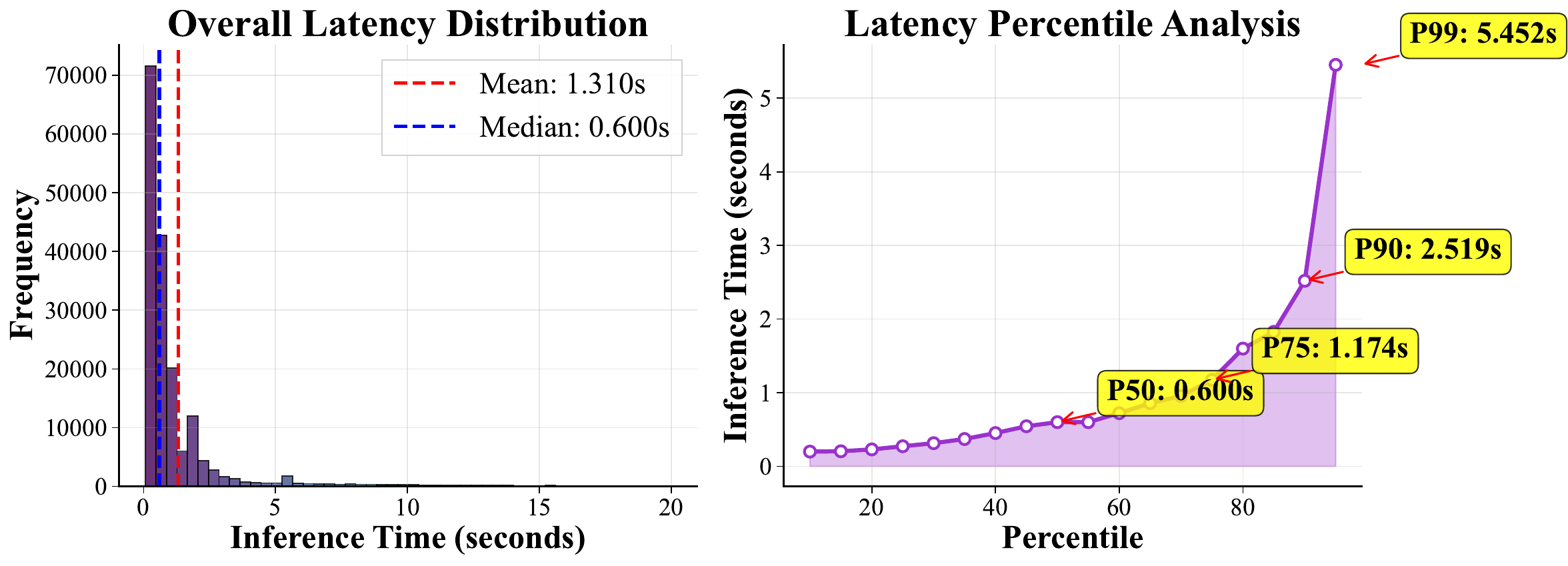}
    \caption{Overall latency histogram (left) and percentile curve (right) across all $\sim$22k model–instance pairs. Dashed lines mark the mean ($\approx$1.31s) and median ($\approx$0.60s); tail extends beyond 5s at P99.}
    \label{fig:rsd-latency-overall}
\end{figure}

\subsubsection{Latency Distribution}
\begin{figure}[h]
    \centering
    \begin{subfigure}{0.49\linewidth}
        \centering
        \includegraphics[width=\linewidth]{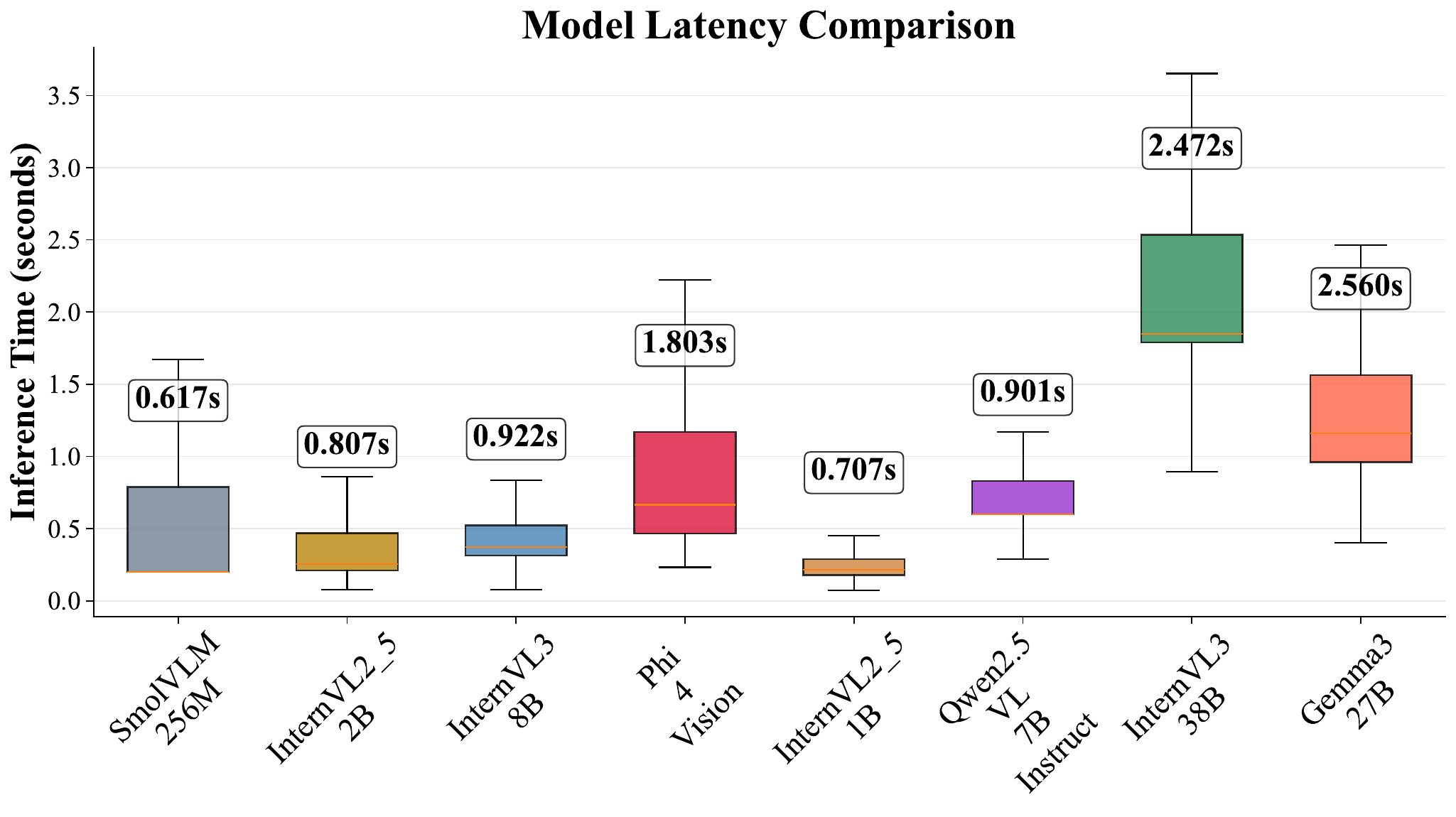}
        \caption{Per-model latency distributions (boxplots with annotated means).}
        \label{fig:rsd-latency-dataset}
    \end{subfigure}\hfill
    \begin{subfigure}{0.49\linewidth}
        \centering
        \includegraphics[width=\linewidth]{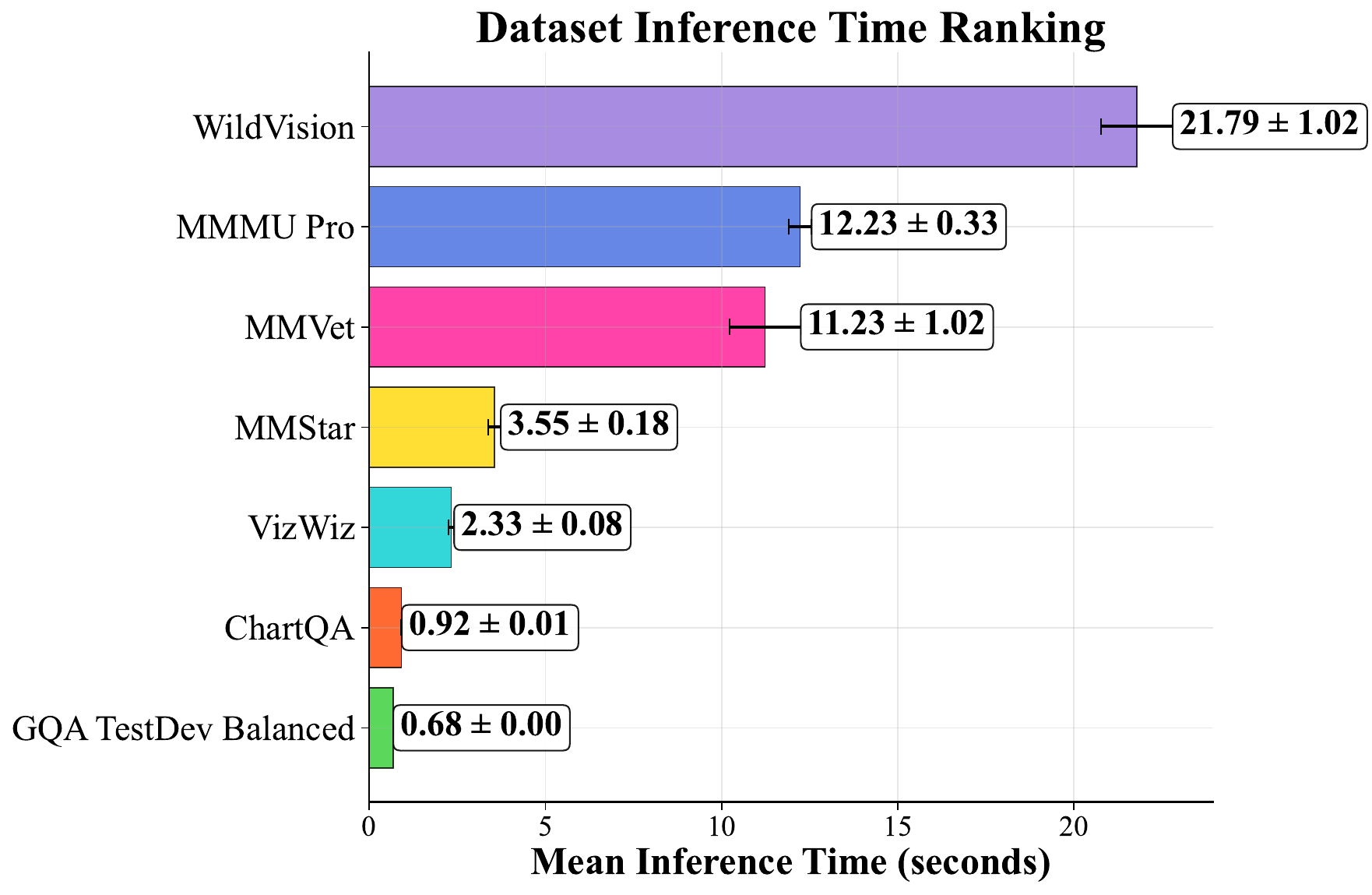}
        \caption{Dataset mean latency (error bars: standard error).}
        \label{fig:rsd-latency-model}
    \end{subfigure}
    \caption{Latency breakdown in RSD: (a)  model-level comparisons on the same instances; (b) dataset-level inference time ranking.}
    \label{fig:rsd-latency-breakdown}
\end{figure}

Latency in RSD is markedly right–skewed (Figure~\ref{fig:rsd-latency-overall}). The median end-to-end time is $\approx\!0.60$s while the mean is $\approx\!1.31$s, indicating a long tail: P75 $\approx\!1.17$s, P90 $\approx\!2.52$s, and P99 $\approx\!5.45$s. Such headroom at the tail motivates scenario-aware routing that prefers $M_{\mathrm{edge}}$ when MES is met. Figure~\ref{fig:rsd-latency-breakdown} decomposes latency by dataset and model. Datasets with heavy OCR or open-world reasoning (WildVision, MMVet, MMMU-Pro) are slowest, whereas structured QA (GQA, ChartQA) is fast. Across models, SVLMs achieve sub-second medians (e.g., SmolVLM-256M~0.62s; InternVL2\_5-1B~0.71s; 2B~0.81s), while LVLMs are slower (InternVL3-38B~2.47s; Gemma3-27B~2.56s). These trends, aligned with Table~\ref{tab:datasets}, provide the latency side of our quality–speed trade-off for routing.

\subsubsection{Analysis Heatmap}
\begin{figure}[h]
    \centering
    \includegraphics[width=1\linewidth]{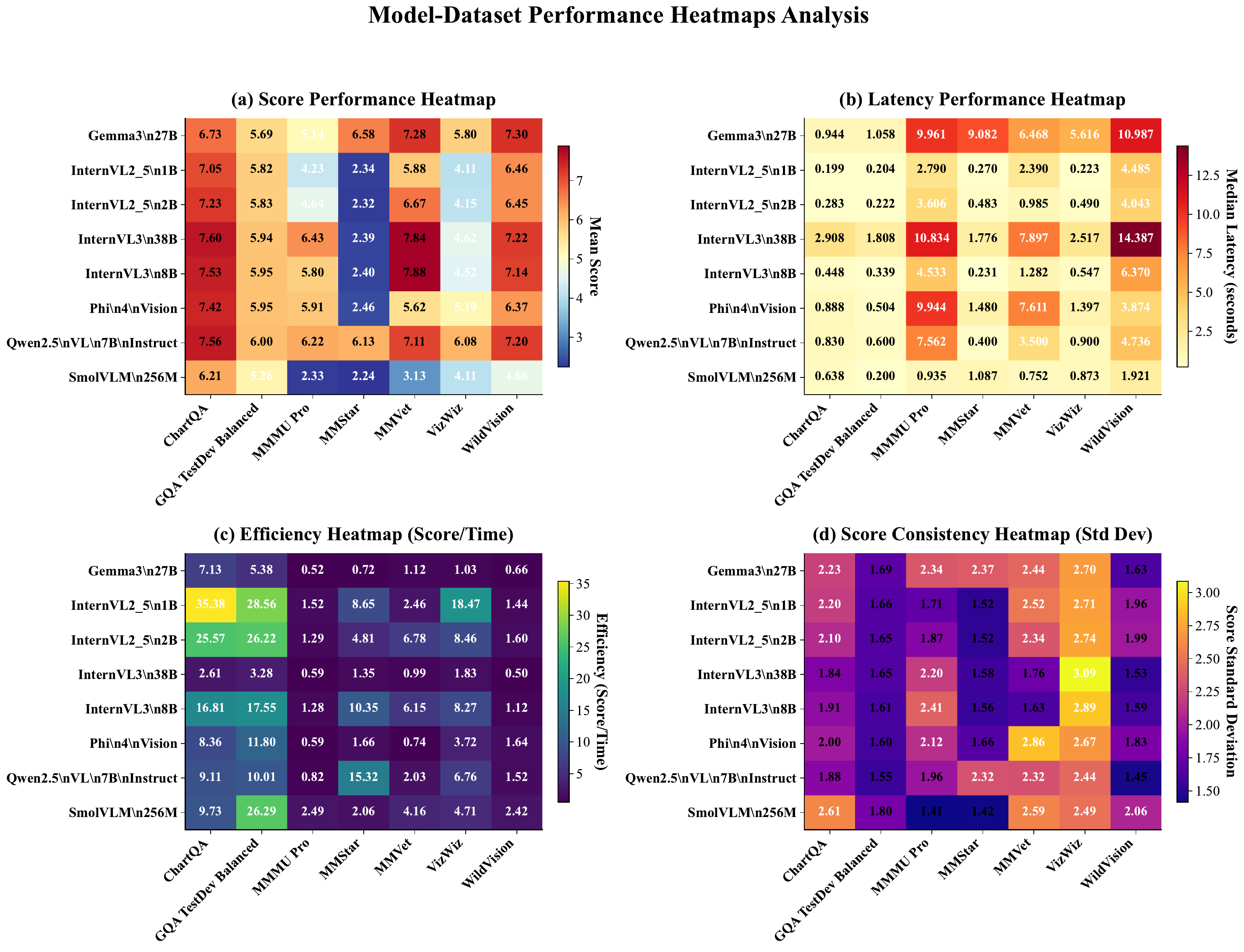}
    \caption{\textbf{Model–Dataset Heatmaps.}
    \textbf{(a)} Score performance (mean score).
    \textbf{(b)} Latency performance (median seconds).
    \textbf{(c)} Efficiency (\emph{score}/\emph{time}).
    \textbf{(d)} Score consistency (standard deviation): a ``hotter'' plasma color (toward bright/purple) indicates larger fluctuation; the cell value is the exact std.\,dev., and white numbers highlight cells with variance \emph{above the global average}.
    \emph{How to read:} scan a \emph{row} to compare one model’s stability across datasets; scan a \emph{column} to compare how challenging a dataset is across models.}
    \label{fig:rsd-heatmaps}
\end{figure}

To examine the accuracy–speed–stability trade–off on the same RSD instances, we summarize per-(model, dataset) statistics in a four-panel heatmap (Figure~\ref{fig:rsd-heatmaps}). Panel (a) shows mean scores (higher is better); (b) shows median latency in seconds (lower is better); (c) reports efficiency as \emph{score/time} (higher is better); and (d) reports score standard deviation (lower is better). Overall, SVLMs attain strong efficiency on structured sets (ChartQA, GQA), while LVLMs deliver higher absolute scores on harder sets (MMVet, MMStar) at a latency cost. WildVision and MMMU-Pro display both large latency and higher variability, matching the difficulty/latency trends reported above.

\section{Additional Experiments}
\label{app:ALL EXPERIMENTS}

\subsection{ADAPTABILITY ACROSS MODELS}
\label{app:adapt}
\begin{figure}[h]
\centering
\includegraphics[width=0.75\linewidth]{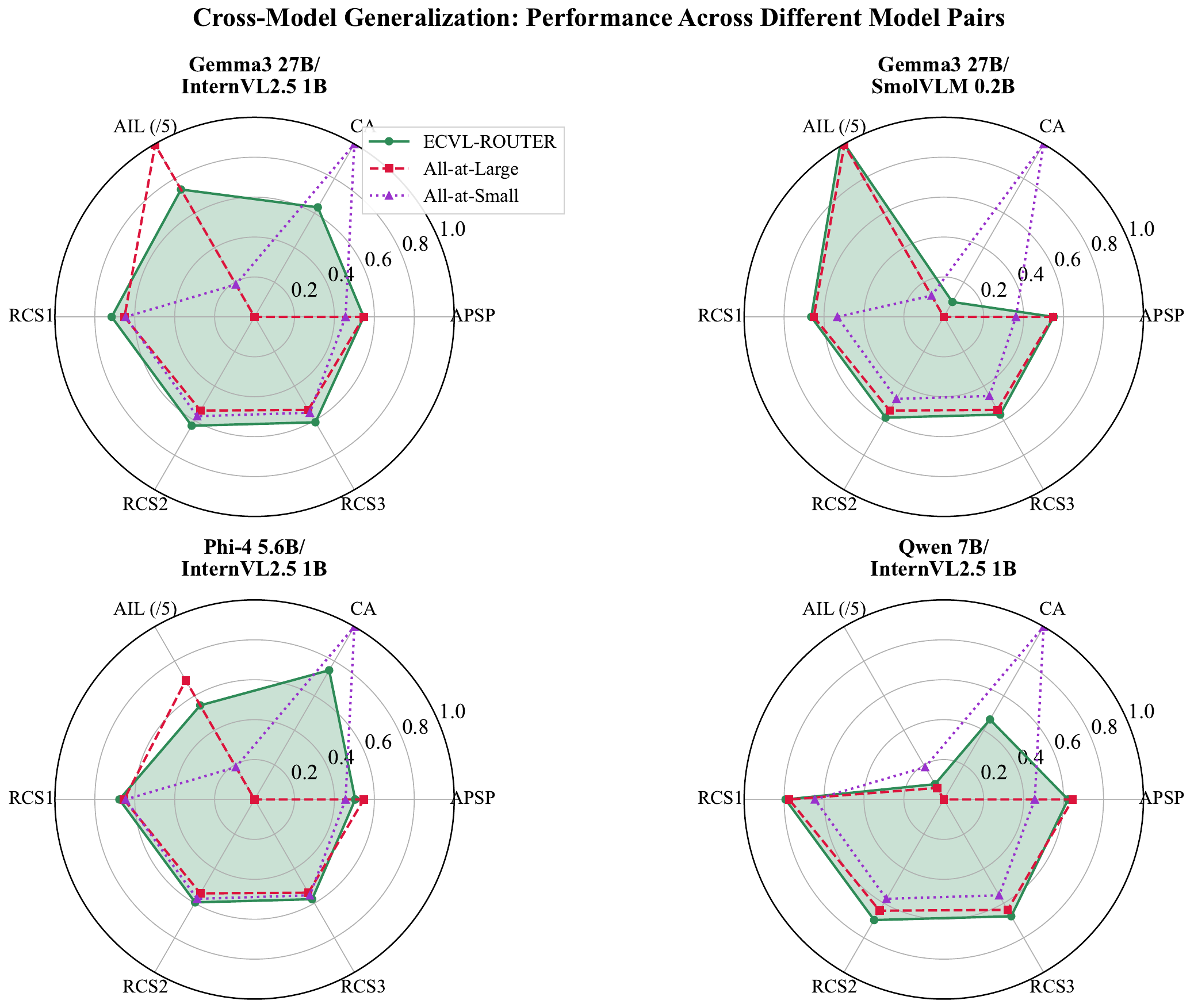}
\caption{\textbf{Cross-model generalization across four model pairs.}
Radar plots compare \emph{ECVL-ROUTER} (green, solid), \emph{All-at-Large} (red, dashed), and \emph{All-at-Small} (purple, dotted) on
\textbf{APSP}(↑), \textbf{RCS1/2/3}(↑), \textbf{CA}(↑, share routed to Small), and \textbf{AIL/5}(↓; latency divided by~5 for scale).
Larger radii indicate better performance for all axes \emph{except} AIL, where larger means slower.
When the Small model is reasonably capable (e.g., pairs with \textbf{InternVL2.5-1B}), ECVL-ROUTER attains higher RCS while keeping latency moderate and CA non-trivial.
With a very weak Small model (e.g., the \textbf{SmolVLM-0.2B} pair), the router intentionally lowers CA (sending most traffic to the large model), yielding limited cost benefit but preserving quality.}
\label{fig:pairs}
\end{figure}
We further supplemented our study with additional model-pair experiments to demonstrate the generality of ECVL-Router. As shown in \autoref{fig:pairs}, when the small model is reasonably strong, ECVL-Router yields a clear improvement in RCS scores—for example, in the case of the Gemma3-27B / InternVL2.5-1B pair. We also experimented with a significantly smaller model, SmolVLM-0.2B, and found that when the performance of the small model is too weak, the model router routes the vast majority of queries to the large model. In such cases, the routing strategy offers little practical benefit in cost savings. Therefore, the smaller model must have sufficient problem-solving capability for the model router to be effective when selecting large-small model pairs.

\subsection{Router Generalization}
\label{app:generate}

\begin{figure}[h]
    \centering
    \includegraphics[width=1\linewidth]{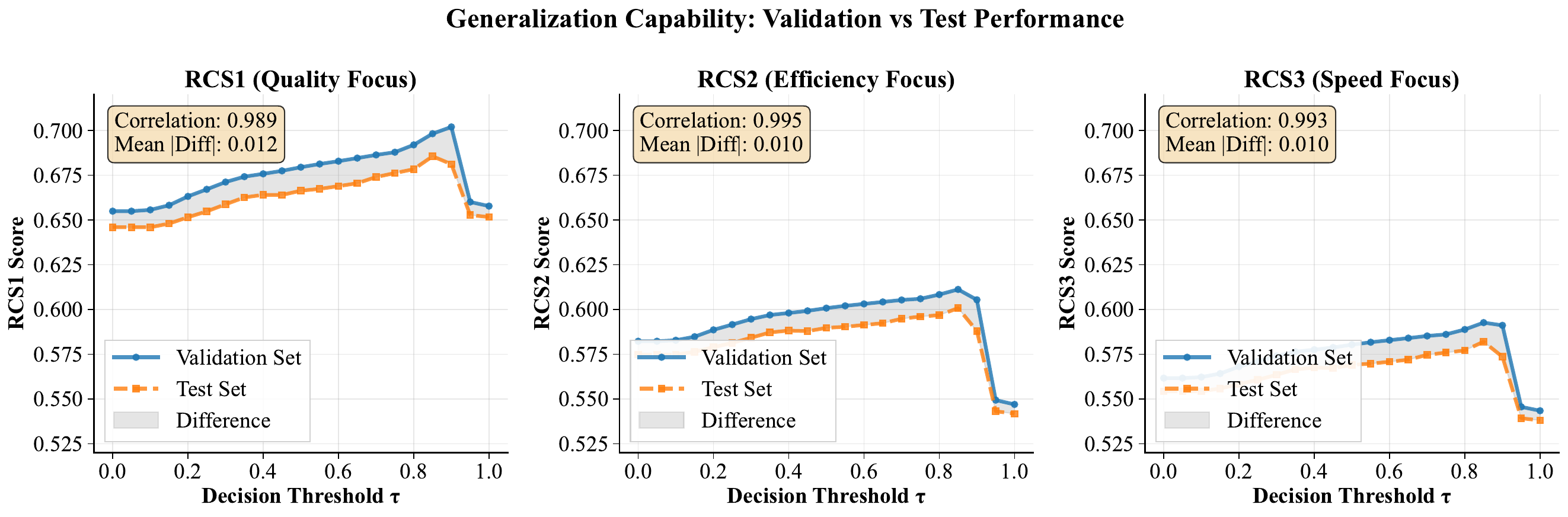}
    \caption{\textbf{Validation--test alignment for selecting the decision threshold $\tau$.}
    RCS1/2/3 vs.\ $\tau$ on the validation set (solid blue) and test set (dashed orange).
    Pearson correlations between the two curves are \emph{0.989} (RCS1), \emph{0.995} (RCS2), and \emph{0.993} (RCS3); mean absolute differences are \emph{0.012}/\emph{0.010}/\emph{0.010}.
    Performance increases with $\tau$ and peaks near $\tau\!\approx\!0.85$--$0.90$, then drops sharply as $\tau$ approaches $1.0$, indicating over-conservative routing and validating grid-search selection on the validation set.}
    \label{fig:align}
\end{figure}

 In this work, the value of $\tau$ is selected via a grid search on the validation set, and the chosen $\tau$ is subsequently used together with the ECVL-Router for the model routing task. The figure \ref{fig:align} illustrates how the RCS on both the validation and test sets vary as $\tau$ changes. It can be observed that the trends of RCS variations on the validation and test sets are largely consistent. The Pearson correlation coefficient between the RCS and the $\tau$ values on the validation and test sets is approximately 0.98, indicating a strong correlation. This strongly supports the validity of using the validation set to determine the optimal $\tau$ value.

\begin{table}[h]
    \caption{Results on cross-domain generalization.}
    \centering
    \begin{tabular}{l l l l l l l}
        \hline
        Model Router & APSP & CA & AIL & RCS1 & RCS2 & RCS3 \\
        \hline
        ECVL-ROUTER & 0.8982 & 0.6316 & 18.02408891 & 1.123 & 0.9547 & 0.9343 \\
        GBDT        & 0.8380 & 0.9582 & 6.100296302  & 1.0953 & 0.9469 & 0.9247 \\
        MLP         & 0.8380 & 0.9758 & 5.119651497  & 1.0981 & 0.9500 & 0.9279 \\
        MF          & 0.8558 & 0.7702 & 7.893076085  & 1.0961 & 0.9403 & 0.9210 \\
        All-at-Large         & 0.9432 & 0.0000 & 24.44174668  & 1.1074 & 0.9188 & 0.9065 \\
        All-at-Small         & 0.8332 & 1.0000 & 4.235969758  & 1.0956 & 0.9490 & 0.9268 \\
        \hline
    \end{tabular}
    \label{tab:cross_domain_generalization}
\end{table}

To evaluate cross-domain generalization, we use InternVL-38B/1B as the model pair. ChartQA, WildVision, MMMU, and MMVet (about 5k samples) serve as the \emph{test} sets, while the rest of RSD is used for training/validation. As summarized in \autoref{tab:cross_domain_generalization}, \textsc{ECVL-Router} attains higher RCS than either All-at-Large or All-at-Small across all three scenarios, demonstrating cross-domain transfer. Compared with in-domain results (Section \ref{subsec:results}), the gains are smaller (roughly $0.5$--$1.5\%$ vs.\ $5$--$6\%$), suggesting routing effectiveness is domain-sensitive; thus, routing strategies and training data should align closely with the intended application domain.


\subsection{online user study}
\label{app:online}
To evaluate the adaptability of our method across different real-world application scenarios, we conducted an online user study involving five participants who interacted with the model router on a dataset of 200 image–text pairs. Two experimental scenarios were designed: 

\textbf{Scenario 1}: Users prioritize response quality, with the MES set to 7, and RCS1 used as the metric in validation stage.

\textbf{Scenario 2}: Users prioritize response speed, with MES set to 5, and RCS3 used as the metric in validation stage.

In each scenario, we trained an ECVL-Router using our routing strategy for online evaluation. For comparison, we also included three baselines: routing all queries to the large model, routing all queries to the small model, and training an ECVL-Router using the win-soft(1) routing strategy.

\begin{figure}[h]
    \centering
    \includegraphics[width=1\linewidth]{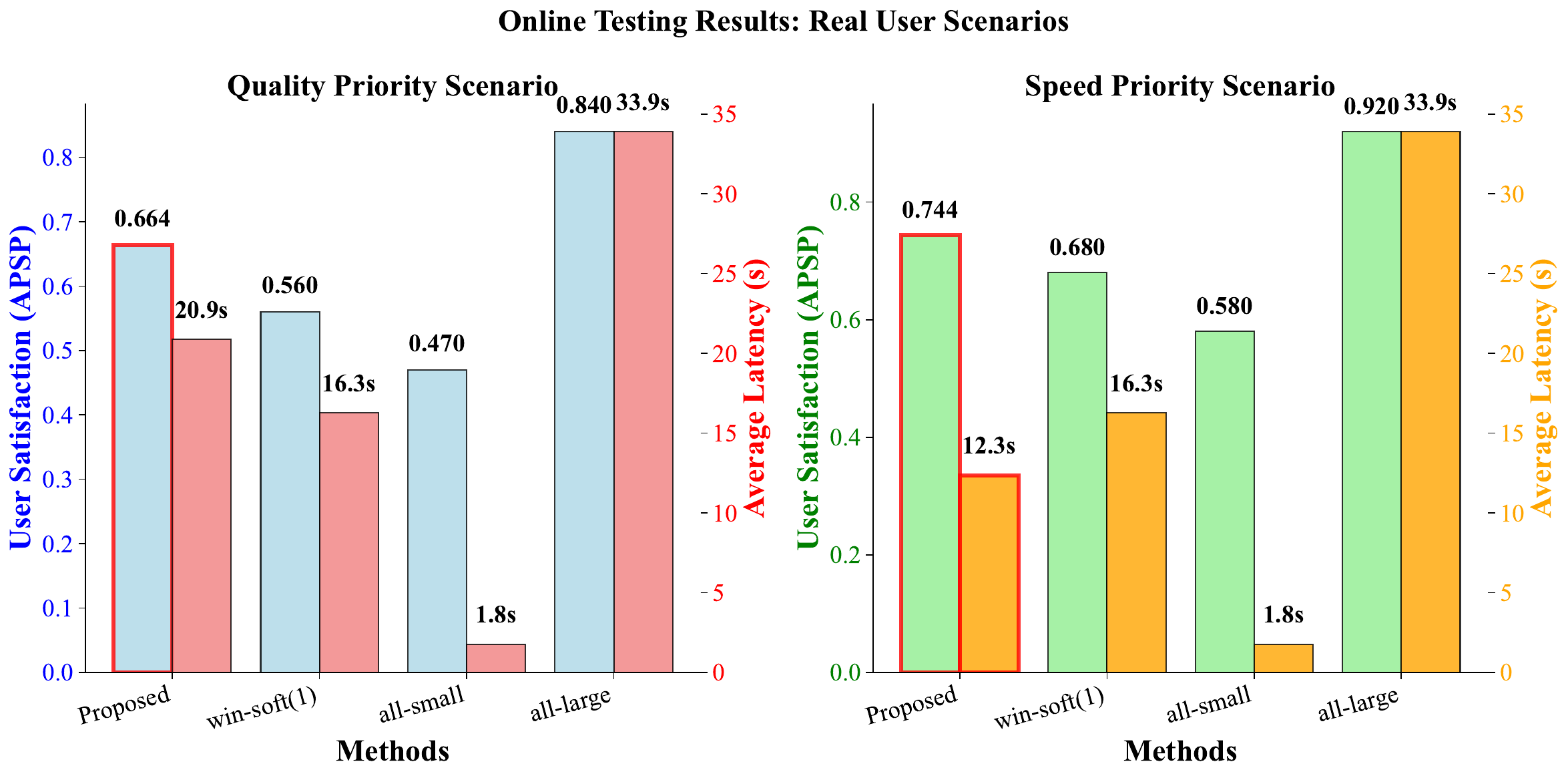}
    \caption{\textbf{Online user study across two scenarios.} APSP (left) and latency (s; right) for four routing policies. Our router adapts to goals: Quality-priority (MES=7/RCS1)—higher APSP than win-soft(1) at moderate latency; Speed-priority (MES=5/RCS3)—higher APSP with lower latency. All-large is highest quality but slowest; all-small fastest yet least satisfying.}
    \label{fig:online_user}
\end{figure}

As shown in \autoref{fig:online_user}, the average APSP and AIL results recorded by the five participants across the two scenarios. In Scenario 1, where response quality is emphasized, our routing strategy achieves higher APSP than win-soft(1), while its AIL is slightly higher. In Scenario 2, where response speed is prioritized, our method outperforms win-soft(1) on both APSP and AIL. This is because our training strategy is designed to maximize user satisfaction rather than to simply favor the stronger model. When MES is set to 5, a larger proportion of queries can already be satisfactorily handled by the small model; as a result, our method successfully routes more queries to the small model while maintaining a high level of user satisfaction.

These online experiments demonstrate that our routing strategy is highly flexible in practice, allowing users to balance response quality and response speed according to the requirements of different application scenarios.

\subsection{Unexpected Query Analysis}
\label{app:unexpect}
Our router tends to favor the small model (SVLM) partly due to the dataset labeling rule: when both the large and small models fail to reach the user’s Minimal Expectation Score (MES), the sample is still labeled as “route to SVLM.” This yields a subset of inherently unsolvable queries in the SVLM bucket.

\begin{figure}[h]
    \centering
    \includegraphics[width=0.9\linewidth]{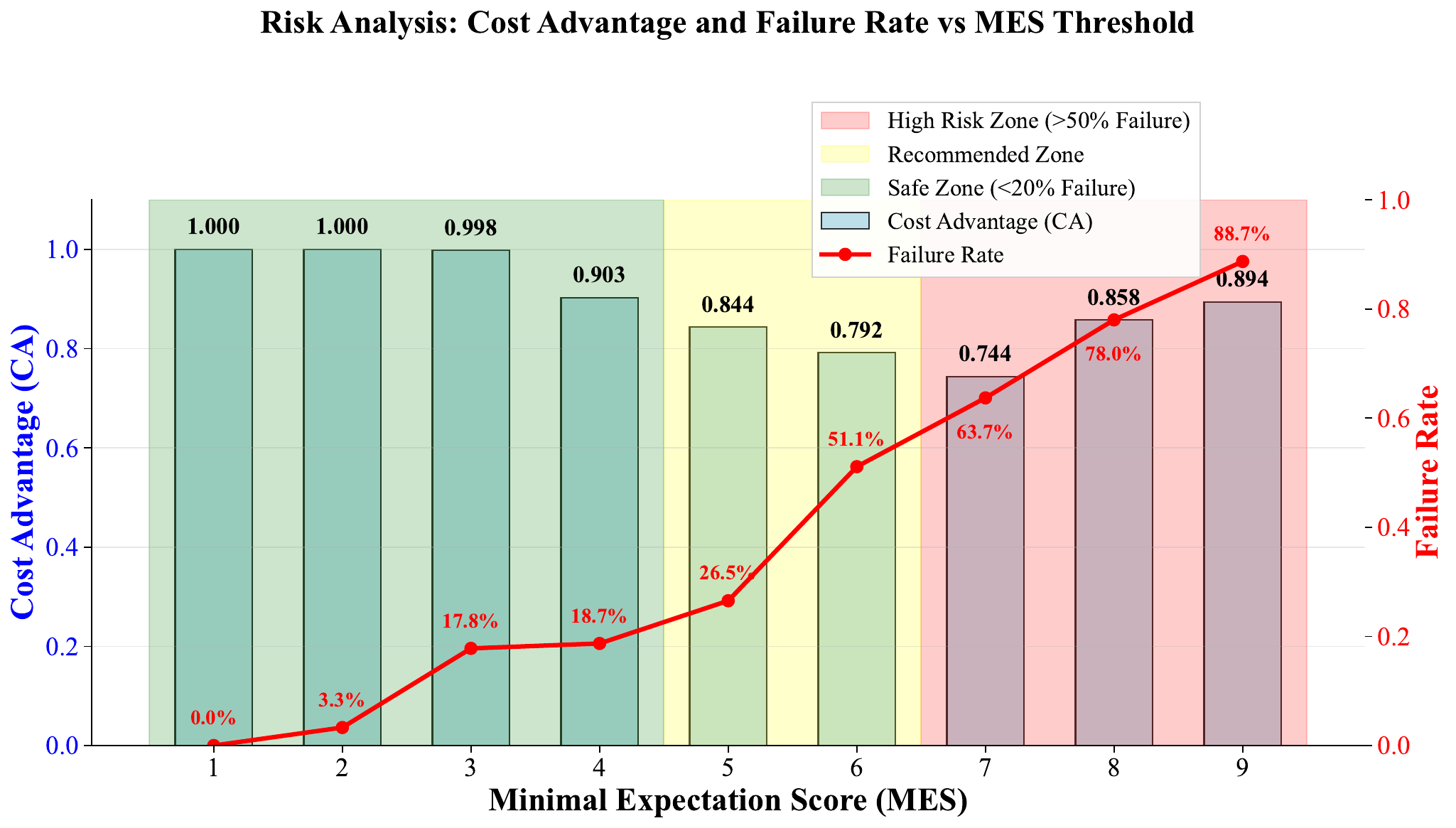}
    \caption{\textbf{MES risk map.} Failure rate (red line, right axis) climbs steeply as MES increases, defining a \emph{safe} region (MES 1–4, $<20\%$ failures), a \emph{caution} region (MES 5–6), and a \emph{high-risk} region (MES $\geq$7, $>50\%$ failures). Cost Advantage (bars, left axis) stays near 1.0 at low MES, bottoms out at MES=7 (0.744), and partially recovers at MES=8–9, illustrating the trade-off between user expectation and routing cost.}

    \label{fig:mes_risk}
\end{figure}

As shown in \autoref{fig:mes_risk}, \textbf{Cost Advantage (CA)} is displayed as \emph{bars on the left $y$-axis}, while the \textbf{failure rate} is the \emph{red line on the right $y$-axis}. In the \emph{safe zone} (MES $=1$–$4$; green band, $<20\%$ failures), most routed queries are solved and CA remains high (1.000/1.000/0.998/0.903). Entering the \emph{caution zone} (MES $=5$–$6$; yellow band), failures rise sharply (26.5\% $\rightarrow$ 51.1\%) and CA drops (0.844 $\rightarrow$ 0.792). In the \emph{high-risk zone} (MES $\geq 7$; red band), failures dominate (63.7\%, 78.0\%, 88.7\%), with CA reaching its minimum at MES $=7$ (0.744) and then partially recovering at MES $=8$–$9$ (0.858/0.894).

\subsection{Additional Plot}
\begin{figure}[H]
    \centering
    \includegraphics[width=1\linewidth]{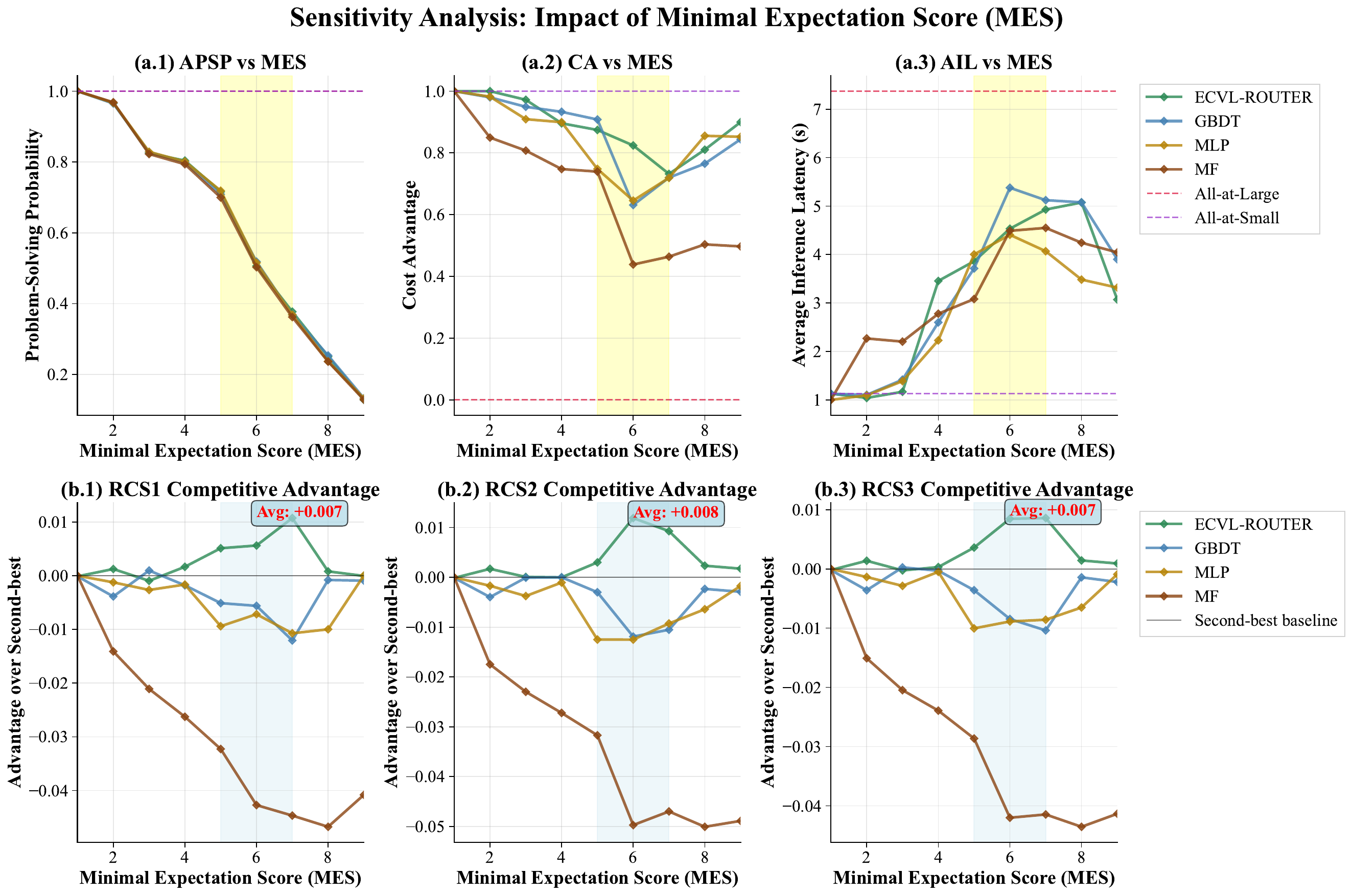}
    \caption{This figure shows the performance trends for all models and routing strategies as the MES value changes.The (a) plots the three metrics performance across different MES, and the (b) panel plots the performance gain of each router relative to the second-best method, as the baseline.}
    \label{fig:sensi-mes}
\end{figure}


\section{Example of Query Hardness and llm judge score}

\begin{figure}[h]
    \centering
    \includegraphics[width=1\linewidth]{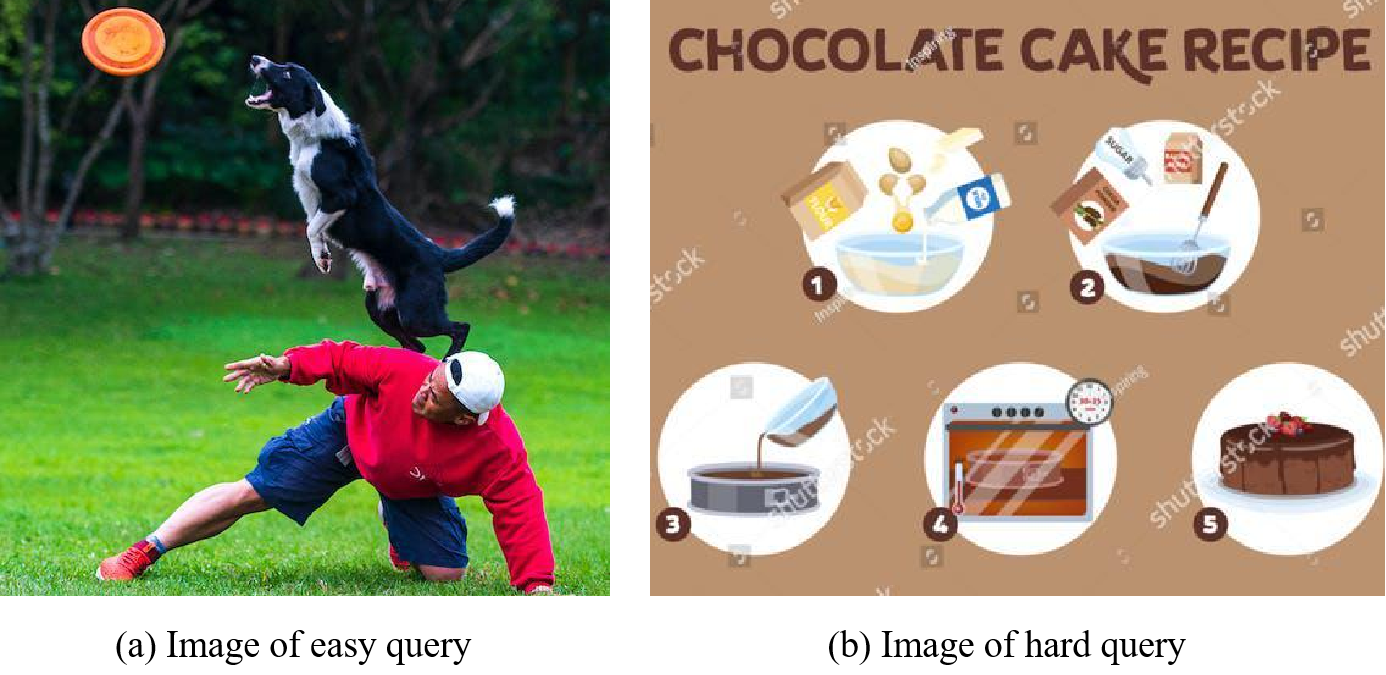}
    \caption{An example of an easy query figure and a hard query figure.}
    \label{fig:query_example}
\end{figure}

\subsection{Easy Query Example}
\textbf{Input Query}: $\langle$ Fig \ref{fig:query_example} (a)$\rangle$ "What is the spatial relation between the frisbee and the man?"

\textbf{InternVL3-38B}:

VLM Response: "The frisbee is in the air, positioned above and slightly in front of the man. The man is crouched on the ground, reaching out towards the frisbee, indicating that he has likely thrown it or is preparing to catch it. The dog is jumping over the man, aiming to catch the frisbee mid-air."

LRJ Reason: "The model answer accurately describes the spatial relation between the frisbee and the man, stating that the frisbee is "in the air, positioned above and slightly in front of the man." This aligns well with the reference answer ("above the man $\langle$OR$\rangle$top left") and provides additional context about the scene, such as the man's posture and the involvement of a dog. The extra details about the man's action and the dog's movement are relevant to understanding the scene but are not strictly necessary for answering the original question, which specifically asked about the spatial relation between the frisbee and the man. The answer is clear, logically structured, and free from errors, but it could be more concise and focused on the direct spatial relationship. Thus, it earns a score of 8 for being accurate, detailed, and well-expressed, though slightly more verbose than required."

LRJ Score: 8

\textbf{InternVL3-8B}:

VLM Response: "The frisbee is in front of the man and slightly above his head, moving in the air towards the left side of the image. The man appears to be throwing it or preparing to catch it. The dog is jumping in the air, aiming to catch the frisbee."

LRJ Reason: "The model answer is mostly accurate and relevant, correctly identifying that the frisbee is above the man and slightly in front of him, which aligns with the reference answer (above the man $\langle$OR$\rangle$ top left). The model also adds contextual details about the man possibly throwing or catching the frisbee and mentions a dog jumping to catch it. While this extra context is not required for the specific spatial relation question, it does not detract from the main answer. However, the answer could be more concise and focused on the direct spatial relation, as the reference answer is. The inclusion of "in front of the man" is a minor deviation, as the primary spatial relation should be "above" or "top left." Overall, the response is accurate and logical but could be improved by focusing more precisely on the spatial relationship asked."

LRJ Score: 7

\textbf{InternVL2.5-1B}:

VLM Response: "The frisbee is in the air, and the man is jumping to catch it."

LRJ Reason: "The model answer is relevant and generally correct in describing the scene: it states that the frisbee is in the air and the man is jumping to catch it. However, it does not directly answer the specific spatial relation question as clearly as the reference answer ("above the man" or "top left"). The response is descriptive but lacks the precise spatial terminology requested. While it gives context, it misses the concise spatial relationship, resulting in a passing but not high score."

LRJ Score: 6

\subsection{Hard Query Example}
\textbf{Input Query}: $\langle$ Fig \ref{fig:query_example} (b)$\rangle$ "Should I add sugar when mixing egg and milk?"

\textbf{InternVL3-38B}:

VLM Response: "No, according to the image, you should add sugar in the second step when you mix it with cocoa powder and other dry ingredients."

LRJ Reason: "The model answer is accurate, relevant, and provides additional helpful context beyond the simple "no" in the reference answer. It correctly states that sugar should not be added when mixing egg and milk, and further clarifies that sugar is added in the second step with cocoa powder and other dry ingredients, presumably based on the referenced image. The explanation is logical and clear, enhancing the user's understanding of the process. However, since the answer relies on information from an image (not shown here), it assumes the user is following a specific recipe, which may not universally apply. For this reason, while the answer is excellent and detailed, it is not fully comprehensive for all contexts, so it does not receive a perfect score."

LRJ Score: 8

\textbf{InternVL3-8B}:

VLM Response: "No, according to the recipe, you should add the sugar during mixing, as shown in step 2."

LRJ Reason: "The model answer is inaccurate and contradicts the reference answer. The user asks whether sugar should be added when mixing egg and milk, and the reference answer is "no." However, the model answer incorrectly states that sugar should be added during mixing, referencing a "step 2" that is not present in the context. This introduces confusion and provides incorrect guidance. While the answer is somewhat relevant to the topic, it lacks accuracy and misleads the user, resulting in a low score. The response does not meet the requirements for correctness or clarity."

LRJ Score: 4

\textbf{InternVL2.5-1B}:

VLM Response: "Yes, you should add sugar when mixing egg and milk. It helps to create a smooth and creamy texture in the chocolate cake."

LRJ Reason: "The model answer is inaccurate and misleading. The user asked whether sugar should be added when mixing egg and milk, and the reference answer is "no." The model incorrectly asserts that sugar should be added and justifies this by mentioning the creation of a smooth and creamy texture in chocolate cake, which is irrelevant to the general question about mixing egg and milk. The response assumes a specific context (chocolate cake) that was not provided by the user and gives advice contrary to the reference answer. The answer lacks accuracy, is not relevant to the general question, and contains significant errors in logic and content."

LRJ Score: 2


\end{document}